\documentclass[final,3p,times]{elsarticle}

\usepackage{amssymb}
\usepackage{amsmath}
\usepackage[T1]{fontenc}
\usepackage{mathtools}         
\usepackage{amsfonts}
\usepackage{mathrsfs}
\usepackage[bold]{hhtensor}
\usepackage{algorithm}
\usepackage{algpseudocode}
\usepackage{enumitem}
\usepackage{booktabs}
\usepackage{threeparttable}

\usepackage{graphicx}
\usepackage{xcolor}
\usepackage{subcaption}
\usepackage{siunitx}
\usepackage{booktabs}
\usepackage{makecell}
\usepackage{multirow}
\usepackage{float}
\usepackage{placeins}
\usepackage{fancyhdr}  
\usepackage{tikz}
\usepackage[nolist]{acronym}

\usepackage{acronym}
\usepackage{pifont}
\usepackage{lineno}
\usepackage{geometry}
\usepackage[colorlinks=true, linkcolor=blue, citecolor=blue, urlcolor=blue]{hyperref}

\begin{acronym}
\acro{PDEs}[PDEs]{partial differential equations}
\acro{BC}{boundary condition}
\acro{KANs}{Kolmogorov-Arnold networks}
\acro{KAN}{Kolmogorov-Arnold network}
\acro{SciML}{Scientific machine learning}
\acro{BCs}{\ac{BC}s}
\acro{MLPs}{multilayer perceptrons}
\acro{ML}{machine learning}
\acro{DNN}{deep neural network}
\acro{RBFs}{radial basis functions}
\acro{PINNs}{physics-informed neural networks}
\acro{DeepONet}{deep operator networks}
\acro{EDNNs}{evolutionary deep neural networks}
\acro{EvoKAN}{evolutionary \ac{KAN}}
\acro{GP}{Gaussian process}
\acro{SAV}{scalar auxiliary variable}
\acro{EDNN}{evolutionary deep neural networks}
\acro{EvoKAN}{evolutionary \ac{KAN}}
\acro{PINN}{physics-informed neural networks}
\end{acronym}

\journal{Computer Methods in Applied Mechanics and Engineering}

\begin{document}
\begin{frontmatter}

\title{Weak-Form Evolutionary Kolmogorov–Arnold Networks
for Solving Partial Differential Equations}

\author[label1]{Bongseok Kim} 
\ead{kim4853@purdue.edu}

\author[label2]{Jiahao Zhang}
\ead{zhan2296@purdue.edu}

\author[label1,label2]{Guang Lin\corref{cor1}}
\cortext[cor1]{Corresponding author}
\ead{guanglin@purdue.edu}

\affiliation[label1]{organization={Purdue University, School of Mechanical Engineering},
            city={West Lafayette},
            postcode={47906}, 
            state={IN},
            country={United States}}

\affiliation[label2]{organization={Purdue University, Department of Mathematics},
            city={West Lafayette},
            postcode={47906},
            state={IN},
            country={United States}}

\begin{abstract}

Partial differential equations (PDEs) form a central component of scientific computing.
Among recent advances in deep learning, evolutionary neural networks have been developed to successively capture the temporal dynamics of time-dependent PDEs via parameter evolution. 
The parameter updates are obtained by solving a linear system derived from the governing equation residuals at each time step.
However, strong-form evolutionary approaches can yield ill-conditioned linear systems due to pointwise residual discretization, and their computational cost scales unfavorably with the number of training samples.
To address these limitations, we propose a weak-form evolutionary Kolmogorov–Arnold Network (KAN) for the scalable and accurate prediction of PDE solutions.
We decouple the linear system size from the number of training samples through the weak formulation, leading to improved scalability compared to strong-form approaches.
We also rigorously enforce boundary conditions by constructing the trial space with boundary-constrained KANs to satisfy Dirichlet and periodic conditions, and by incorporating derivative boundary conditions directly into the weak formulation for Neumann conditions.
In conclusion, the proposed weak-form evolutionary KAN framework provides a stable and scalable approach for PDEs and contributes to scientific machine learning with potential relevance to future engineering applications.

\end{abstract}

\begin{graphicalabstract}
\includegraphics[width=1\linewidth]{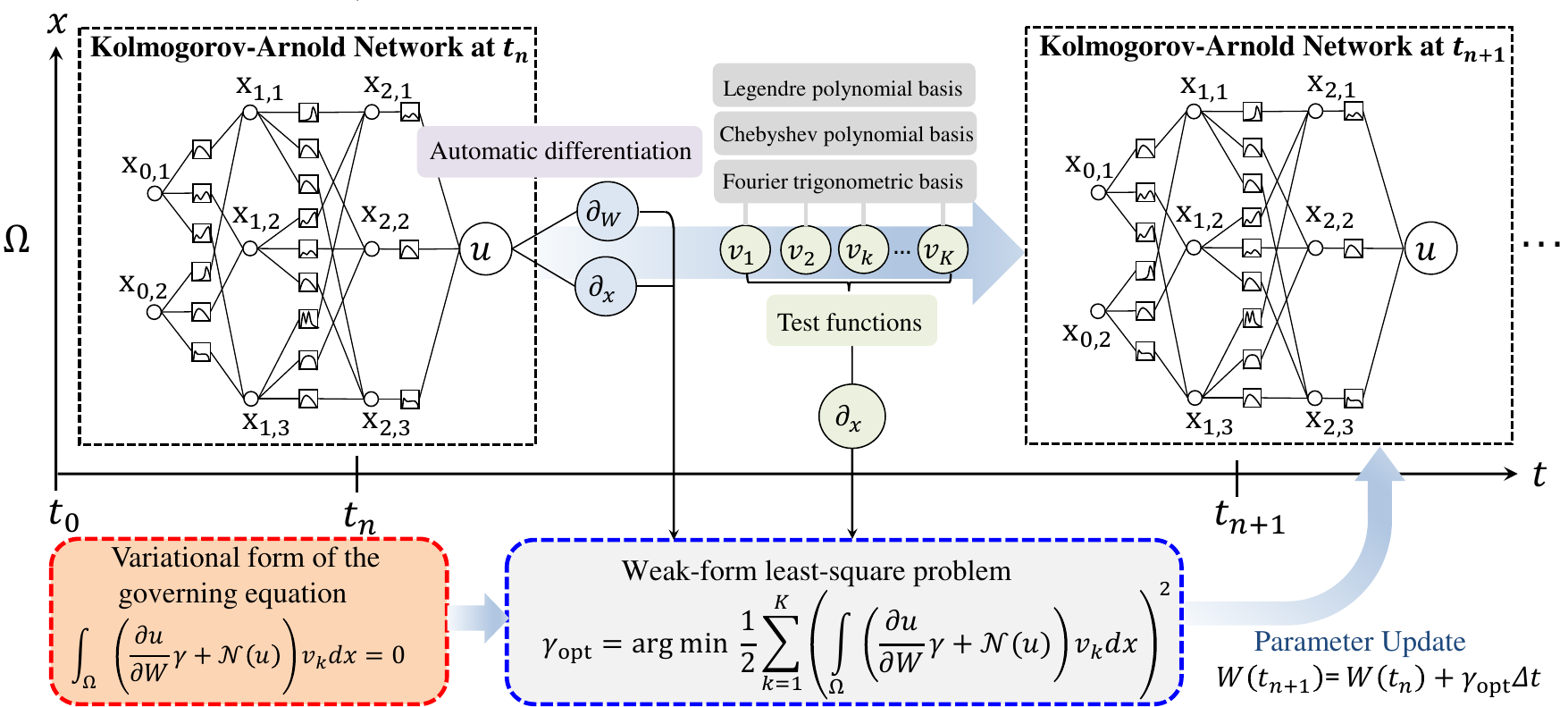}
\end{graphicalabstract}

\begin{highlights}
\item A weak-form evolutionary PDE framework improving stability and scalability.
\item Fixed-size, well-conditioned parameter updates independent of collocation resolution.
\item Dirichlet, periodic, and Neumann boundary conditions consistently incorporated.
\item Improved computational efficiency over strong-form approaches.
\end{highlights}


\end{frontmatter}

\section{Introduction}

Deep learning has been explored for approximating solution operators of \ac{PDEs} as well as for representing nonlinear mappings and complex solution manifolds~\cite{yu2018deep,weinan2020machine,han2018solving}, forming an important line of work within scientific machine learning (SciML).
Approaches such as the deep Galerkin method (DGM)~\cite{sirignano2018dgm} and physics-informed neural networks (PINNs)~\cite{karniadakis2021physics, cai2021physics, raissi2019physics} formulate PDE learning as a strong-form residual minimization problem and predict the solution of the underlying \ac{PDEs}.
As an extension of PINNs, subsequent extensions have aimed to improve stability, convergence, and the ability to resolve multi-scale features.
For instance, adaptive sampling and curriculum strategies~\cite{wu2023comprehensive, mao2023physics} redistribute collocation points toward regions with large residuals or sharp gradients, which mitigates training failures on stiff or high-frequency solutions.
Domain-decomposition approaches~\cite{jagtap2020xpinn, shukla2021parallel}, including conservative PINNs and extended PINNs, partition the space--time domain into subdomains that can be trained in parallel and better handle complex geometries and discontinuities.
Multi-scale architectures that incorporate Fourier features~\cite{wang2021eigenpinn} or multiple frequency components~\cite{cai2019mscalednn} help reduce the spectral bias of standard PINNs and improve accuracy for highly oscillatory solutions.
Other extensions address nonlocal operators and fractional dynamics, such as the fractional PINN~\cite{pang2019fpinn} for space--time fractional advection--diffusion equations.

The deep learning approach has also progressed toward learning solution operators directly, leading to frameworks such as DeepONet~\cite{lu2021deeponet}, the Fourier neural operator (FNO)~\cite{li2021fourier,kovachki2021neural}, and the Laplace neural operator (LNO)~\cite{cao2023laplacian}.  
Further studies have introduced neural-operator surrogates for PDEs, including methods based on wavelet representations~\cite{garg2024_vswno} and continuous space--time operator learning~\cite{chen2025_ndo}.  
These approaches demonstrate that neural networks can provide mesh-free and generalizable approximations for forward, inverse, and parametric PDE problems.

Along with advances in neural network approaches for solving PDEs, Kolmogorov--Arnold networks (KANs)~\cite{liu2024kan,liu2024kan2}, inspired by the classical representation theorem of Kolmogorov and Arnold~\cite{kolmogorov1957representation,kolmogorov1961representation,braun2009constructive}, have been examined as an alternative to standard multilayer perceptrons (MLPs) in scientific machine learning.
While recent works have clarified strengths of MLP-based models, these networks still depend on preset activation functions combined with trainable linear parameters~\cite{apicella2021survey,trentin2001networks}.  
KANs depart from this structure by replacing fixed activations with trainable spline functions, allowing the network to adjust its nonlinearities locally and to represent features at multiple resolutions~\cite{liu2024kan,liu2024kan2}.
This construction offers advantages in interpretability and provides additional flexibility in representing multivariate mappings. 
Several studies have explored KAN-based formulations in the context of partial differential equations. 
Liu et al.~\cite{liu2024kan} embedded KANs within the PINN framework~\cite{raissi2019physics} for a two-dimensional Poisson equation, 
while Abueida et al.~\cite{abueidda2024deepokan} introduced DeepOKAN, an operator-learning formulation based on radial-basis 
representations and evaluated it on two-dimensional orthotropic elasticity and transient diffusion problems. 
Wang et al.~\cite{wang2025kolmogorov} examined KANs across strong-, energy-, and inverse-form PDE formulations within the PIKAN/KINN framework. 
In addition, Thakolkaran et al.~\cite{thakolkaran2025can} proposed input-convex KAN architectures for hyperelastic constitutive modeling.

Beyond architectural design, another perspective considers how network parameters change when the dynamics vary in time. Models with fixed parameters typically approximate the space--time solution map in one training stage by adding time as an input variable.
An alternative perspective is to update the network in a time-marching manner so that the parameters evolve with the temporal dynamics.
For instance, Du et al.~\cite{du2021evolutional} introduced this idea through the evolutionary deep neural network (EDNN), where the trainable parameters advance according to a time-discretized form of the governing PDE. Their method first trains a baseline network and then propagates its parameters forward in time without repeated optimization. 
Zhang et al.~\cite{zhang2024energy} extended this paradigm to operator learning by applying similar time-marching updates to neural operators for parameterized PDEs. 
More recent efforts incorporated KAN-based representations into evolutionary schemes. Lin et al.~\cite{lin2025energy} introduced energy-dissipative KANs for solving complex PDE problems, and Kim et al.~\cite{kim2025bekan} used Gaussian radial basis functions within the KAN ansatz to enforce boundary constraints during evolution.
Despite these developments, existing evolutionary approaches predominantly rely on the strong form of the governing equations, leaving weak-form evolutionary formulations underexplored.

Unlike strong form approaches that enforce PDEs through pointwise residual minimization, weak formulations offer three main advantages:
(i) integration by parts reduces the differential order and weakens the regularity requirements, enabling more accurate approximation of discontinuous or steep gradient solutions~\cite{de2024wpinns,wang2025wf};
(ii) the formulation avoids reliance on high order automatic differentiation, improving numerical stability~\cite{xu2021weak};
and (iii) replacing pointwise collocation constraints with quadrature based integral constraints allows the PDE to be enforced globally with far fewer evaluation points~\cite{xu2021weak}.
Within this line of work, Kharazmi et al.~\cite{kharazmi2019variational} introduced the variational physics informed neural network (VPINN) based on a Petrov Galerkin formulation, where the trial space is a neural network and the test space is constructed from classical basis functions. 
Related extensions, including variational formulations incorporated into operator learning frameworks~\cite{goswami2022physics,berrone2023enforcing}, have reported improved robustness for multi-scale or highly irregular PDE solutions.
However, despite these developments, no existing study has applied a variational formulation to evolutionary neural networks.

Inspired by variational approaches for solving PDEs, we propose a weak-form evolutionary framework that incorporates a weak residual formulation into an evolutionary network. 
We represent the solution on a nonlinear manifold parameterized by boundary-constrained KANs and select polynomial or trigonometric functions as test functions.
Specifically, we employ a KAN architecture with Gaussian radial basis functions and impose hard constraints to satisfy the Dirichlet or periodic boundary conditions.
To solve the PDE problems, we update the network parameters by solving a weak-form least-squares problem derived from the weak residual of the governing equation, where trigonometric or polynomial test functions define the test space. 
In contrast to least-squares formulations based on strong forms, the resulting linear system consists of integral terms evaluated over the computational domain, which mitigates pointwise sensitivity, improves numerical stability, and yields more reliable solutions for PDEs with steep or discontinuous features.
At each time step, we solve this integral system and advance the network parameters using a forward Euler update.
Compared with prior evolutionary network approaches utilizing the strong form, the proposed method provides several advantages:
\begin{enumerate}
    \item \textbf{Scalable computational efficiency} --- The proposed weak-form framework decouples the parameter update system size from the number of training samples and fixes it with respect to the number of test functions, resulting in improved scalability compared to strong-form approaches.
    
    \item \textbf{Improved conditioning and accuracy }--- By reducing the order of differentiation and replacing pointwise residual evaluation with weak residual projections, the proposed method yields well-conditioned linear systems and maintains accuracy for solutions approaching discontinuities.

    \item \textbf{Rigorous enforcement of boundary conditions} --- The proposed method rigorously enforces all types of boundary conditions: Dirichlet and periodic conditions are imposed through boundary-constrained trial spaces, while Neumann conditions are incorporated directly into the weak formulation via integration by parts.
\end{enumerate}

This paper is organized as follows.
Sections~\ref{sec:KANs} and~\ref{sec:KANs_RBFs} introduce the network architectures employed in this work.
Section~\ref{sec:EvoKAN_RBFs} presents the formulation of the evolutionary Kolmogorov--Arnold network.
Section~\ref{sec:VEKAN} describes the proposed weak-form evolutionary approach.
Section~\ref{Sec:numerical_experiments} reports numerical results for representative PDE problems to assess the performance of the method.
Finally, Section~\ref{sec:conclusion} concludes the study and outlines potential future research directions.

\section{Methodology}
\label{sec:EvoKAN}

In this section, we introduce the \ac{KAN}, the \ac{KAN} enhanced with \ac{RBFs}, and the evolutionary \ac{KAN} formulated in the strong form. We then present the proposed weak-form evolutionary \ac{KAN}, which reformulates the evolutionary approach in the weak form.

\subsection{Kolmogorov–Arnold networks}
\label{sec:KANs}

\begin{figure}[htp!]

    \centering
    \begin{subfigure}[b]{0.34\linewidth}
        \centering
        \includegraphics[width=\linewidth]{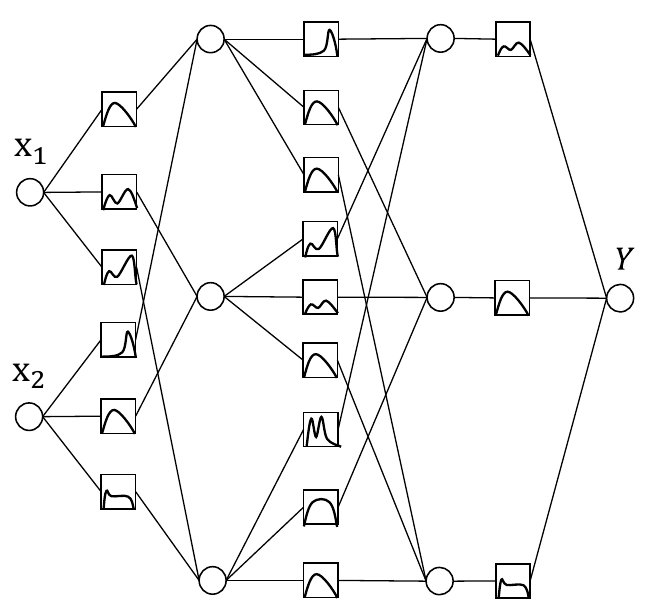}
        \subcaption{KAN structure}
    \end{subfigure}
    \begin{subfigure}[b]{0.37\linewidth}
        \centering
        \includegraphics[width=\linewidth]{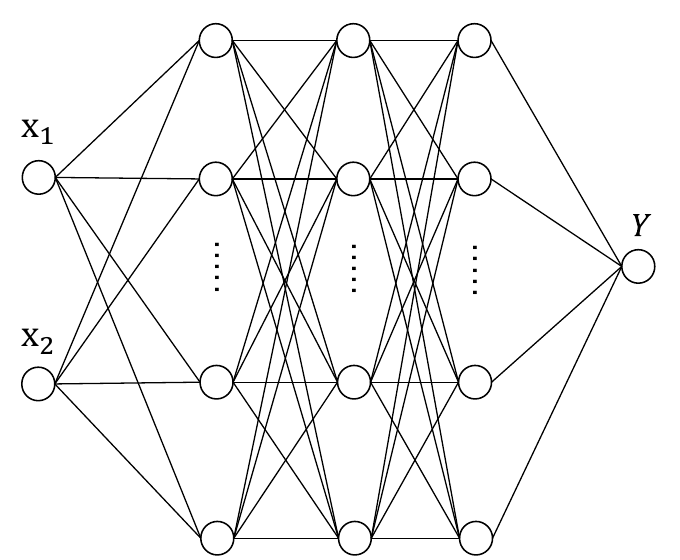}
        \subcaption{DNN structure}
    \end{subfigure}
\caption{
Comparison between the Kolmogorov--Arnold Network (KAN) and a conventional fully connected DNN.
(a) The KAN architecture employs functional mappings on each edge, enabling richer
nonlinear representations and adaptive feature extraction.
(b) The DNN architecture is composed of fully connected layers with activation functions.}
\label{fig:KAN_DNN}
\end{figure}

\ac{KANs} illustrated in Fig.~\ref{fig:KAN_DNN}(a) place a trainable univariate function on every edge linking two neurons and force each neuron to actively combine transformed edge inputs to produce an output~\cite{liu2024kan, liu2024kan2}.
In comparision,
the multilayer perceptrons in a \ac{DNN} in Fig.~\ref{fig:KAN_DNN}(b) impose a single activation function on each layer~\cite{sze2017efficient}, whereas \ac{KANs} distribute nonlinear transformations across all connections and thereby yield enhanced expressiveness and interpretability~\cite{fakhoury2022exsplinet}.  
The Kolmogorov--Arnold theorem establishes a foundation for this architecture by proving that continuous multivariate mappings on bounded domains can be represented through finite sums and compositions of continuous one-dimensional functions~\cite{koppen2002training,lin1993realization,lai2021kolmogorov,leni2013kolmogorov,he2023optimal}.
Based on the Kolmogorov--Arnold theorem, a smooth mapping \( f : [0,1]^n \to \mathbb{R} \) admits the representation
\begin{equation}
    f(\mathbf{x})
    = \sum_{q=1}^{2n+1}
      \Phi_q\!\left(
        \sum_{p=1}^n \phi_{q,p}(x_p)
      \right),
    \label{eq:KART}
\end{equation}
where each inner mapping \( \phi_{q,p} : [0,1] \to \mathbb{R} \) modifies a single coordinate and each outer mapping \( \Phi_q : \mathbb{R} \to \mathbb{R} \) merges all contributions.  
All mappings satisfy continuity conditions imposed by the theorem.

Univariate components in Eq.~\eqref{eq:KART} use third-order B\text{-}spline basis functions~\cite{liu2024kan}.  
The inner stage organizes all one-dimensional mappings into the collection $\mathbf{\Phi}=\{\phi_{ij}\}$ and each element acts on a single coordinate while producing one intermediate channel.  
The Kolmogorov--Arnold representation fixes the dimensional structure because the first functional stage receives an input of size $n$ and generates $2n+1$ channels and this requirement yields $n_{\rm in}=n$ and $n_{\rm out}=2n+1$.  
A final reduction stage compresses the $2n+1$ intermediary channels into one scalar through an additional univariate mapping and this step imposes $n_{\rm out}=1$ at the terminal stage.

To generalize beyond the canonical $n$-input and $2n+1$-channel structure,
this mechanism constructs a hierarchical system of scalar compositions
that mirrors the functional representation introduced in
Eq.~\eqref{eq:KART}.
A sequence of layers is specified by the list $[n_0,n_1,\ldots,n_L]$,
where $n_\ell$ denotes the number of neurons in layer $\ell$.
Each pair of adjacent layers admits one univariate mapping for every directed connection, and the notation
\begin{equation}
    \phi_{\ell,j,i},
    \quad
    \ell=0,\ldots,L-1,
    \quad
    i=1,\ldots,n_\ell,
    \quad
    j=1,\ldots,n_{\ell+1},
\end{equation}
indexes all such transformations.  
Let $x_{\ell,i}$ denote the value produced by the $i$th neuron in layer $\ell$. Then every map consumes $x_{\ell,i}$ and outputs the quantity
\begin{equation}
\label{eq:phi_lji}
    \tilde{x}_{\ell,j,i}=\phi_{\ell,j,i}(x_{\ell,i}),
\end{equation}
and neuron $j$ in layer $\ell+1$ receives the aggregated input
\begin{equation}
    x_{\ell+1,j}=\sum_{i=1}^{n_\ell}\tilde{x}_{\ell,j,i}
               =\sum_{i=1}^{n_\ell}\phi_{\ell,j,i}(x_{\ell,i}) ,
    \quad
    j=1,\ldots,n_{\ell+1}.
\end{equation}
The transformations in one layer can be collected into a matrix-valued operator 
\begin{equation}
    \mathbf{\Phi}_\ell=
    \begin{pmatrix}
        \phi_{\ell,1,1}(\cdot) & \phi_{\ell,1,2}(\cdot) & \cdots & \phi_{\ell,1,n_\ell}(\cdot) \\
        \phi_{\ell,2,1}(\cdot) & \phi_{\ell,2,2}(\cdot) & \cdots & \phi_{\ell,2,n_\ell}(\cdot) \\
        \vdots & \vdots & & \vdots \\
        \phi_{\ell,n_{\ell+1},1}(\cdot) & \phi_{\ell,n_{\ell+1},2}(\cdot) & \cdots & \phi_{\ell,n_{\ell+1},n_\ell}(\cdot)
    \end{pmatrix},
\end{equation}
which acts on the vector $\mathbf{x}_\ell$ through composition, yielding
\begin{equation}
    \mathbf{x}_{\ell+1}=\mathbf{\Phi}_\ell\mathbf{x}_\ell .
\end{equation}
A full \ac{KAN} applies $L$ such operators in succession so the network output becomes
\begin{equation}
    \text{KAN}(\mathbf{x})
    =
    (\mathbf{\Phi}_{L-1}\circ\mathbf{\Phi}_{L-2}
     \circ\cdots\circ\mathbf{\Phi}_1\circ\mathbf{\Phi}_0)\mathbf{x}.
\end{equation}
When the final layer contains a single neuron with $n_L=1$, the network defines a scalar mapping $f(\mathbf{x})$ and the repeated compositions expand into the following expression:
\begin{equation}
\begin{aligned}
    f(\mathbf{x})
    &=\sum_{i_{L-1}=1}^{n_{L-1}}
      \phi_{L-1,i_L,i_{L-1}}
      \Bigg(
          \sum_{i_{L-2}=1}^{n_{L-2}}
          \phi_{L-2,i_{L-1},i_{L-2}}
          \Bigg(
                \cdots
                \sum_{i_2=1}^{n_2}
                \phi_{2,i_3,i_2}
                \Big(
                    \sum_{i_1=1}^{n_1}
                    \phi_{1,i_2,i_1}
                    \big(
                        \sum_{i_0=1}^{n_0}\phi_{0,i_1,i_0}(x_{i_0})
                    \big)
                \Big) \cdots
          \Bigg)
      \Bigg).
\end{aligned}
\end{equation}

\subsection{Kolmogorov-Arnold networks with radial basis functions}
\label{sec:KANs_RBFs}

A \ac{KAN} architecture can adopt \ac{RBFs} to streamline the construction of univariate transformations and to accelerate numerical operations~\cite{li2405kolmogorov}.  
The \ac{RBFs} evaluate a scalar input through a function that depends only on the distance from a fixed center~\cite{orr1996introduction, buhmann2000radial}.
In particular, the \ac{RBFs} build an approximation of a target mapping by combining multiple radially symmetric components that concentrate their influence near selected points in the input domain.  
Incorporating this idea into the \ac{KAN} framework leads to a representation in which every univariate transformation $\phi_{l,j,i}$ in Eq.~\eqref{eq:phi_lji} is replaced by an RBF expansion of the form
\begin{equation}
\label{eq:RBFs_approximation}
\hat{\phi}_{l,j,i}(x)=\sum_{k=1}^{g} w^{k}_{l,j,i}\,\psi(\lVert x-c_{k}\rVert),
\end{equation}
and this approximation satisfies
\begin{equation}
\phi_{l,j,i}(x)\approx\hat{\phi}_{l,j,i}(x).
\end{equation}
The coefficients $w^{k}_{l,j,i}$ determine the contribution of each radial component and the points $c_{k}$ specify the centers that define the radial distances inside $\psi$.  
All layers with indices $l=0,\ldots,L-1$ employ this formulation and every connection from neuron $i$ in layer $l$ to neuron $j$ in layer $l+1$ adopts the same expansion strategy.


In Eq.~\eqref{eq:RBFs_approximation}, the formulation employs Gaussian functions as the radial components, which gives the kernel
\begin{equation}
    \psi(r)=\exp\!\left(-\frac{r^{2}}{2h^{2}}\right).
\end{equation}
The argument \(r\) measures the distance from an evaluation point to a specified center, and the parameter \(h\) regulates the width and influence of the kernel.  
Gaussian RBFs, after suitable linear transformations, reproduce sequences of cubic B\,-spline basis functions~\cite{li2405kolmogorov}.

Replacing each univariate mapping in the original construction with an RBF expansion produces an alternative sequence of layer operators.  
Every operator in this modified sequence uses \ac{RBFs}  $\hat{\phi}_{l,j,i}$ in place of $\phi_{l,j,i}$ and the collection of these approximations forms the transformed layer operators $\mathbf{\Psi}_{l}$.  
Applying all such operators in succession defines an RBF-based variant of the architecture and the resulting transformation acting on an input vector $\mathbf{x}$ is given by
\begin{equation}
\label{eq:RadialKAN}
    \text{RBF-KAN}(\mathbf{x})
    =
    \left(
        \mathbf{\Psi}_{L-1}
        \circ
        \mathbf{\Psi}_{L-2}
        \circ
        \cdots
        \circ
        \mathbf{\Psi}_{1}
        \circ
        \mathbf{\Psi}_{0}
    \right)\mathbf{x}.
\end{equation}
A single-neuron final layer with $n_{L}=1$ converts the network into a scalar-valued mapping.  
In this situation the output becomes $f(\mathbf{x})$ and the nested structure expands into
\begin{equation}
\begin{aligned}
    f(\mathbf{x})
    &=
      \sum_{i_{L-1}=1}^{n_{L-1}}
      \hat{\phi}_{L-1,i_{L},i_{L-1}}
      \Bigg(
         \sum_{i_{L-2}=1}^{n_{L-2}}
         \hat{\phi}_{L-2,i_{L-1},i_{L-2}}
         \Bigg(
                 \cdots
                 \sum_{i_{2}=1}^{n_{2}}
                 \hat{\phi}_{2,i_{3},i_{2}}
                 \Big(
                     \sum_{i_{1}=1}^{n_{1}}
                     \hat{\phi}_{1,i_{2},i_{1}}
                     \big(
                         \sum_{i_{0}=1}^{n_{0}}
                         \hat{\phi}_{0,i_{1},i_{0}}(x_{i_{0}})
                     \big)
                 \Big) \cdots
         \Bigg)
      \Bigg).
\end{aligned}
\end{equation}


\subsection{Evolutionary Kolmogorov--Arnold networks: Strong-form}
\label{sec:EvoKAN_RBFs}

\begin{figure}[htp!]
    \centering
        \begin{subfigure}[b]{0.9\linewidth}
        \centering
        \includegraphics[width=\linewidth]{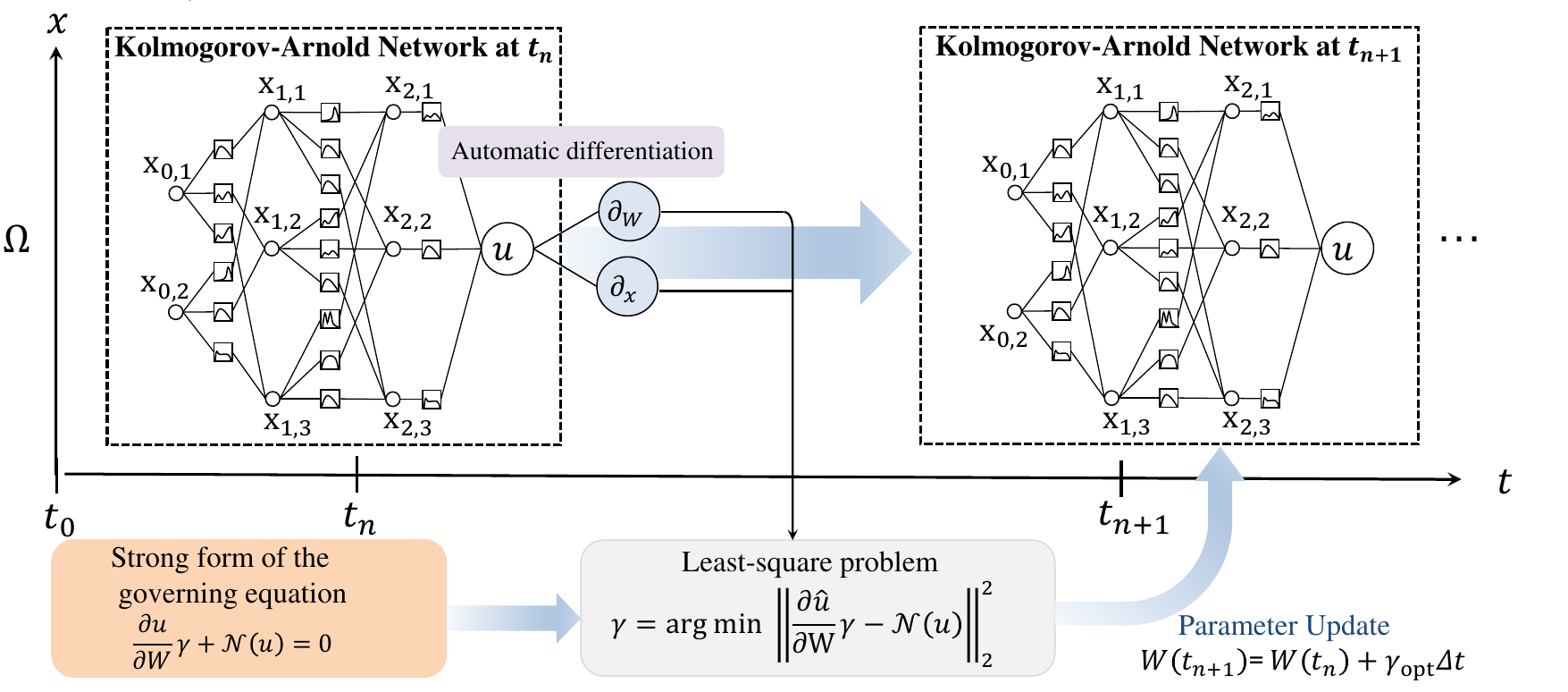}
    \end{subfigure} 
\caption{
Evolutionary Kolmogorov--Arnold network with Gaussian \ac{RBFs}: Strong-form formulation.  
At each time step, the parameters ${W}(t)$ evolve by minimizing the PDE residual functional
$\displaystyle 
\mathcal{J}(\gamma)=\tfrac{1}{2}\!\int_{\Omega}\!\big\|\tfrac{\partial \hat{u}}{\partial {W}}\gamma+\mathcal{N}(\hat{u})\big\|_2^2\,\mathrm{d}\mathbf{x}$,
where the optimal update direction satisfies 
$\displaystyle 
\frac{\partial {W}}{\partial t}=\arg\min_{\gamma}\mathcal{J}(\gamma)$.
After obtaining $\gamma_{\text{opt}}$, the network parameters are explicitly updated in time using the forward Euler scheme
$\displaystyle 
{W}(t_{n+1}) = {W}(t_n) + \gamma_{\text{opt}}\Delta t$,
yielding a discrete-time parameter evolution associated with the strong-form PDE residual minimization.
}
    \label{fig:EvoRadialKAN}
\end{figure}

This section generalizes the RBF-KAN framework to incorporate an evolutionary learning mechanism~\cite{du2021evolutional, zhang2024energy, lin2025energy} to capture the temporal dynamics of \ac{PDEs}. As a starting point, we consider a general form of a nonlinear PDE accompanied by an initial condition:
\begin{equation}
\label{eqn:general-pde}
 \begin{aligned}
    &\frac{\partial u}{\partial t} + \mathcal{N}(u) = 0, \\
    &u(\mathbf{x}, 0) = f(\mathbf{x}),
 \end{aligned}
\end{equation}
In Eq.~\eqref{eqn:general-pde}, the function $u(\mathbf{x}, t) = (u_1, u_2, \dots, u_m)$ represents a multicomponent field, where $\mathbf{x} = (x_1, x_2, \dots, x_d)$ denotes the spatial coordinates, and $\mathcal{N}$ indicates a nonlinear differential operator acting on $u$.

We now represent the solution $u$ using the RBF-KAN approximation $\hat{u}$, parameterized by a network with $L+1$ layers:
\begin{equation}
\hat{u}(\mathbf{x}, {W}(t)) = (\hat{u}_1, \hat{u}_2, \dots, \hat{u}_m) = \text{RBF-KAN}(\mathbf{x}),
\end{equation}
where ${W}(t)$ is a time-dependent vector that collects all trainable parameters of the network.
Applying the chain rule yields the following expression for the time derivative of $\hat{u}$:
\begin{equation}
\frac{\partial \hat{u}}{\partial t} = \frac{\partial \hat{u}}{\partial W} \frac{\partial {W}}{\partial t},
\end{equation}
where the derivative $\frac{\partial {W}}{\partial t}$ governs the direction of parameter evolution.  
In the evolutionary network, we require the derivative $\frac{\partial {W}}{\partial t}$ at each time step.  
For this purpose, we solve the following optimization problem, where we minimize $\mathcal{J}$ derived from the residual of Eq.~\eqref{eqn:general-pde}:
\begin{equation}\label{Wmin}
\frac{\partial {W}}{\partial t} = \arg\min_{\gamma} \mathcal{J}(\gamma), \quad \mathcal{J}(\gamma) = \frac{1}{2} \int_{X} \left\Vert \frac{\partial \hat{u}}{\partial {W}} \gamma + \mathcal{N}(\hat{u}) \right\Vert_2^2 \, \mathrm{d}\mathbf{x}.
\end{equation}
By the first-order optimality condition, we seek the optimal solution of Eq.~\eqref{Wmin} by solving the following system:
\begin{equation}\label{OptCond}
\nabla_{\gamma} \mathcal{J}(\gamma_{\text{opt}}) = 
\int_{\Omega} 
\left( \frac{\partial \hat{u}}{\partial {W}}\right)^T
\left(
\frac{\partial \hat{u}}{\partial {W}}  \gamma_{\text{opt}} + \mathcal{N}(\hat{u}) \right)\mathrm{d} \mathbf{x} = 0.
\end{equation}
To approximate the solution $\gamma_{\text{opt}}$ to Eq.~\eqref{OptCond}, we recast Eq.~\eqref{OptCond} into a least-squares formulation as follows:
\begin{equation}\label{OptCondApprox}
\mathbf{J}^{T} \mathbf{J} \hat{\gamma}_{\text{opt}} + \mathbf{J}^{T} \mathbf{N}=0,
\end{equation}
Here, $\mathbf{J}$ indicates the sensitivity matrix of the network prediction with respect to trainable parameters, whereas $\mathbf{N}$ denotes the residual values obtained by evaluating the governing equation at selected collocation nodes. The entries of these matrices are defined as follows:
\begin{equation}
\left( \mathbf{J} \right)_{ij} = \frac{\partial \hat{u}^i}{\partial {W}_j}, \quad \left( \mathbf{N} \right)_i = \mathcal{N}(\hat{u}^i),
\end{equation}
where the index $i = 1, 2, \dots, N_{\hat{u}}$ refers to the evaluation locations and $j = 1, 2, \dots, N_W$ labels the trainable parameters of the network.
The entries of both $\mathbf{J}$ and $\mathbf{N}$ are computed using automatic differentiation. After computing $\gamma_{\text{opt}}$, we update the network parameters using forward Euler method:
\begin{equation}
    {W}(t_{n+1}) = {W}(t_n) + \gamma_{\text{opt}} \, \Delta t.
\end{equation}

We summarize the evolutionary \ac{KANs} with Gaussian \ac{RBFs} in Fig.~\ref{fig:EvoRadialKAN} under the strong-form formulation, and delineate the implementation process in Algorithm~\ref{alg:evoKAN_strong}.
The network parameters are updated over time in the direction $\gamma$ derived from the governing PDE, enabling the model to reflect the time-dependent behavior of the solution.
Each evolved network state corresponds to a solution snapshot at a given time, and continued updates yield the full solution trajectory.

\begin{algorithm}[h!]
\caption{Evolutionary Kolmogorov--Arnold Network: Strong-form}
\label{alg:evoKAN_strong}
\begin{algorithmic}[1]
\small
\Require Nonlinear operator $\mathcal{N}$; initial condition $f(\mathbf{x})$; collocation points $\{\mathbf{x}^i\}_{i=1}^{N_{\hat{u}}} \subset X$
\State \textbf{Given:} Time step $\Delta t$, number of time steps $M$, number of collocation points $N_{\hat{u}}$

\State $t_0 \gets 0$
\State Initialize $W(t_0)$ such that $\hat{u}(\mathbf{x},W(t_0)) \approx f(\mathbf{x})$

\For{$n = 0,1,\ldots,M-1$}

    \State Evaluate $\hat{u}^i = \hat{u}(\mathbf{x}^i, W(t_n))$ and store $\hat{u}^i$ together with $W(t_n)$
    \State Compute residual entries $N_i^{(n)} = \mathcal{N}(\hat{u}^i)$ and construct the residual vector $N^{(n)}$
    \State Compute sensitivity entries $J_{ij}^{(n)} = \dfrac{\partial \hat{u}^i}{\partial W_j}$ and construct the sensitivity matrix $J^{(n)}$
    \State Solve $J^{(n)\mathsf{T}} J^{(n)} \gamma_{\mathrm{opt}}^{(n)} = -\,J^{(n)\mathsf{T}} N^{(n)}$ for $\gamma_{\mathrm{opt}}^{(n)}$
    \State Update $W(t_{n+1}) \gets W(t_n) + \Delta t\,\gamma_{\mathrm{opt}}^{(n)}$

\EndFor

\State \Return $\{W(t_n)\}_{n=0}^{M}$ and $\{\hat{u}(\mathbf{x},W(t_n))\}_{n=0}^{M}$

\end{algorithmic}
\end{algorithm}

\subsection{Evolutionary Kolmogorov--Arnold networks: Weak-form}
\label{sec:VEKAN}

\subsubsection{Trial space: Boundary-constrained KAN}
\label{sec:BC}

To construct a trial space that satisfies boundary conditions by network
design, the boundary constraints are incorporated directly into the basis
functions of the evolutionary KAN.
With this construction, all admissible functions generated by the trial space inherently satisfy the Dirichlet and periodic boundary conditions, independent of the trainable parameters. Neumann boundary conditions are enforced during the evolutionary stage by incorporating them directly into the weak-form.
Further details on boundary-constrained KAN architectures can be found
in~\cite{kim2025bekan}.

For homogeneous Dirichlet boundaries, the KAN architecture employs two
types of scaling functions.  
At the first hidden layer, each basis function constructed from Gaussian
\ac{RBFs} is multiplied by a factor that vanishes on the
boundary.  
A representative choice is
\begin{equation}
h_1(\mathbf{x})
=
\prod_{i=1}^{d} (x_i - k_i)^{p_i},
\qquad 
0 < p_i \le 1,
\end{equation}
and this factor forces every basis response to be zero whenever
\(\mathbf{x}\) lies on the domain boundary.
In the subsequent layers, activations are rescaled by a second map that
preserves zero values,
\begin{equation}
h_2(z) = z^{\,q},
\qquad
0 < q \le 1.
\end{equation}
Once a zero value appears at some layer, later layers maintain the zero
value under this scaling, and the prediction \(\hat{u}(\mathbf{x})\) therefore
satisfies the homogeneous Dirichlet condition on the boundary.
In the case of non-homogeneous boundary conditions, a lifting function \(l(\mathbf{x},t)\) is introduced, leading to the following trial function:
\begin{equation}
u(\mathbf{x};W(t))
=
\hat{u}(\mathbf{x};W(t))
+
l(\mathbf{x},t).
\end{equation}

For periodic boundary conditions, the spatial coordinates are first mapped to a 
periodic feature representation.
For instance, a one-dimensional coordinate \(x\) can be embedded using multiple Fourier components,
\begin{equation}
x \mapsto 
\left(
\sin(\omega x),\,
\cos(\omega x),\,
\sin(2\omega x),\,
\cos(2\omega x),\,
\dots
\right),
\qquad
\omega = \frac{2\pi}{L},
\end{equation}
and the number of harmonics is selected according to the spectral content of
the target function.  
The periodic feature representation satisfies the boundary identity, and any
function produced by composing these features with the KAN layers preserves
finite-order periodicity,
\begin{equation}
u^{(\ell)}(a) = u^{(\ell)}(b),
\qquad
0 \le \ell \le k,
\end{equation}
where \(k\) corresponds to the highest spatial derivative order in the PDE.
For multidimensional problems, each coordinate is independently embedded
through the same periodic mapping, and the resulting feature vectors are
concatenated prior to evaluation by the radial basis layers.  

\subsubsection{Weak-form evolutionary Kolmogorov--Arnold networks}

\begin{figure}[htp!]
\centering
    \begin{subfigure}[b]{0.9\linewidth}
        \includegraphics[width=\linewidth]{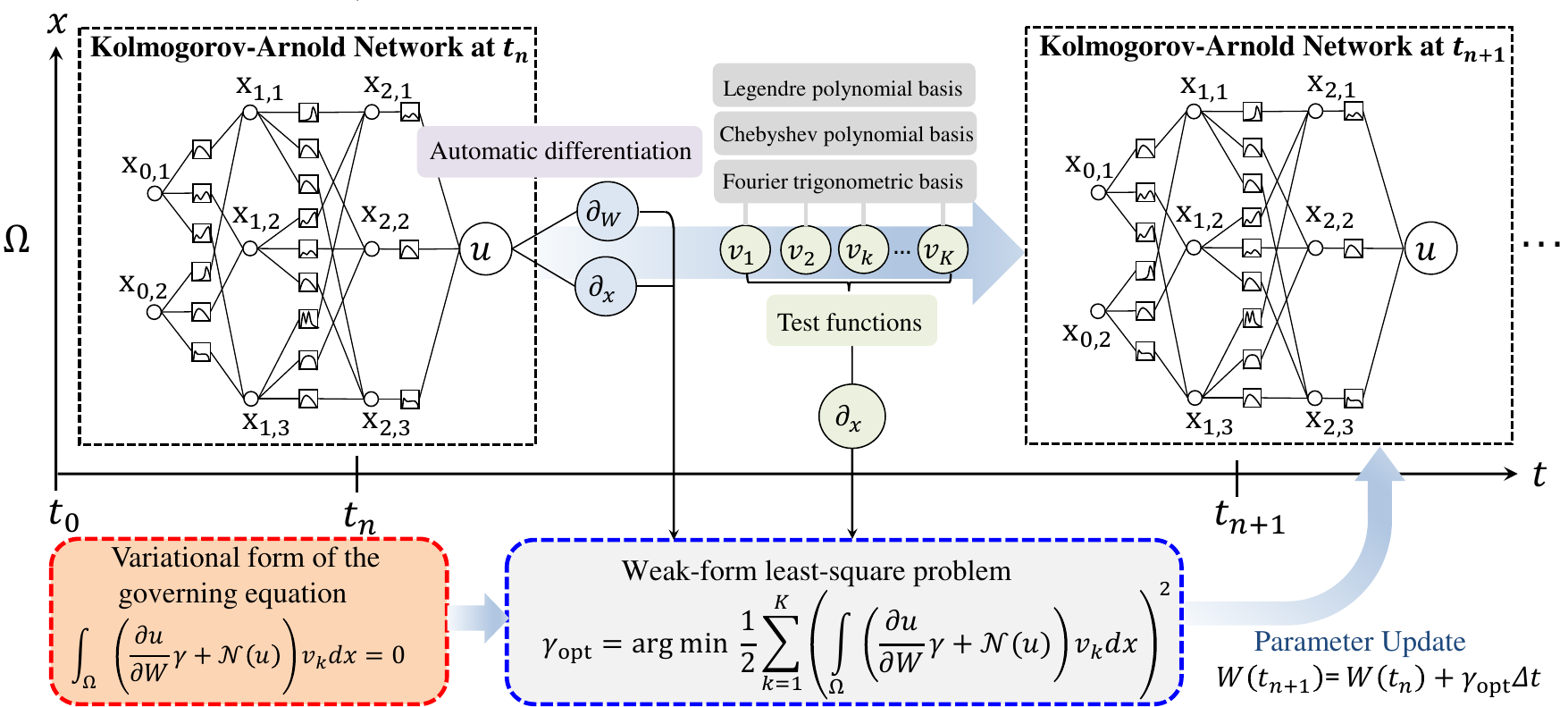}
    \end{subfigure}
\caption{
Weak-form evolutionary Kolmogorov--Arnold network.  
At each time step, the parameters $W(t)$ are updated by minimizing the weak--form residual functional
$\mathcal{J}(\gamma) = \tfrac{1}{2}\sum_{k=1}^{K}\!\left(\int_{\Omega}\big(\tfrac{\partial \hat{u}}{\partial W}\gamma + \mathcal{N}(\hat{u})\big)v_k(\mathbf{x})\,\mathrm{d}\mathbf{x}\right)^{2}$,
where $\{v_k\}_{k=1}^{K}$ denote the test functions.
The optimal update direction is given by $\gamma_{\text{opt}} = \arg\min_{\gamma}\mathcal{J}(\gamma)$,
and the network parameters are advanced in time via the forward Euler step $W(t_{n+1}) = W(t_n) + \gamma_{\text{opt}}\Delta t$,
resulting in a discrete-time parameter evolution associated with the weak-form residual minimization.
}
    \label{fig:VEKAN}
\end{figure}

\begin{algorithm}[h!]
\caption{Evolutionary Kolmogorov-Arnold Network: Weak-form}
\label{alg:weak_evo_kan}
\begin{algorithmic}[1]
\small
\Require Nonlinear operator $\mathcal{N}$; test functions $\{v_k\}_{k=1}^K$; domain $\Omega$; quadrature nodes $\{x_i,w_i\}_{i=1}^{N_x}$; initial condition $u_0(x)$
\State \textbf{Given:} time step $\Delta t$, number of time steps $M$, parameter dimension $P$

\State Initialize $W(t_0)$ such that $u(x,W(t_0)) \approx u_0(x)$

\For{$n = 0,1,\ldots,M-1$}

    \State Evaluate $u^i = \hat{u}(x_i, W(t_n))$ and store $\hat{u}^i$ together with $W(t_n)$
    \State Compute weak-form sensitivity entries 
           $J_{kj} = \sum_{i=1}^{N_x} w_i \, \dfrac{\partial \hat{u}}{\partial W_j}(x_i) v_k(x_i)$
    \State Compute weak-form residual entries 
           $N_k = \sum_{i=1}^{N_x} w_i \, \mathcal{N}(\hat{u}(x_i)) v_k(x_i)$
    \State Assemble $J^{(n)}$ from the rows $(J_{k1},\ldots,J_{kP})$
    \State Assemble $N^{(n)}$ from entries $N_k$
    \State Solve $J^{(n)\mathsf{T}} J^{(n)} \gamma_{\mathrm{opt}}^{(n)}
                  = -\,J^{(n)\mathsf{T}} N^{(n)}$
    \State Update $W(t_{n+1}) \gets W(t_n) + \Delta t\,\gamma_{\mathrm{opt}}^{(n)}$

\EndFor

\State \Return $\{W(t_n)\}_{n=0}^{M}$ and $\{u(\mathbf{x},W(t_n))\}_{n=0}^{M}$

\end{algorithmic}
\end{algorithm}

A weak-form evolutionary \ac{KAN} framework approximates the parameter evolution of time-dependent \ac{PDEs} by enforcing the governing equation in an integral sense.  
The overall structure of this weak-form update process is summarized
schematically in Fig.~\ref{fig:VEKAN}.
Consider a nonlinear evolution problem of the form
\begin{equation}
\frac{\partial u}{\partial t}(x,t) + \mathcal{N}(u(x,t)) = 0,
\qquad
u(x,0) = u_0(x),
\end{equation}
where $t \in [0,T]$, $x \in \Omega \subset \mathbb{R}^d$, and $u(x,t)$ denotes the unknown state field.  
The operator $\mathcal{N}$ represents a nonlinear differential operator in space acting on $u$, and $u_0(x)$ specifies the prescribed initial condition at $t=0$.

In the evolutionary network setting, the solution $u$ is approximated by a parametric model $\hat{u}(x, W(t))$, where the parameter vector $W(t)\in\mathbb{R}^P$ evolves in time.
The temporal evolution of $W(t)$ is described by an auxiliary vector $\gamma$ through the relation
\begin{equation}
\frac{\partial \hat{u}}{\partial t}(x,t)
=
\frac{\partial \hat{u}}{\partial W}(x,W(t))\,\gamma,
\end{equation}
and substitution into the governing equation defines the residual
\begin{equation}
R(x;W,\gamma)
=
\frac{\partial \hat{u}}{\partial W}(x,W)\,\gamma
+
\mathcal{N}(\hat{u}(x,W)).
\end{equation}
To construct a weak formulation, a finite collection of test functions 
$\{v_k\}_{k=1}^{K}$ on $\Omega$ is introduced.  
In all experiments, we employ low-order polynomial or trigonometric bases whose
cardinality is chosen to adequately resolve the dominant spatial modes of the solution.
Empirically, we observe that once $K$ exceeds the effective modal complexity of
the solution, further increases in $K$ lead to only marginal improvements.
For each $v_k$, the residual is projected in a weak sense by integrating over the spatial domain:
\begin{equation}
\int_{\Omega} R(x;W,\gamma)\,v_k(x)\,dx
=
\int_{\Omega}
\left(
\frac{\partial u}{\partial W}(x,W)\,\gamma
+
\mathcal{N}(u(x,W))
\right)
v_k(x)\,dx
\simeq 0,
\qquad
k = 1,\ldots,K.
\end{equation}
These conditions enforce the PDE in a weak sense along the directions spanned by the test functions, which can reduce the influence of high-frequency components in the residual.

The direction $\gamma$ is determined by minimizing a least-squares functional that aggregates the weak residuals over the entire test set.  
Specifically, the optimal update direction is obtained as
\begin{equation}
\gamma_{\text{opt}}
=
\arg\min_{\gamma} J(\gamma),
\qquad
J(\gamma)
=
\frac12
\sum_{k=1}^{K}
\left(
\int_{\Omega}
\left(
\frac{\partial u}{\partial W}(x,W)\,\gamma
+
\mathcal{N}(u(x,W))
\right)
v_k(x)\,dx
\right)^{2}.
\end{equation}
Each term in the sum represents the squared weak residual associated with a single test function, and the functional $J(\gamma)$ quantifies the overall weak residual.
The first-order optimality condition for this minimization problem requires the gradient of $J$ with respect to $\gamma$ to vanish.
The gradient can be written as
\begin{equation}
\nabla_{\gamma} J(\gamma)
=
\sum_{k=1}^{K}
\left[
\int_{\Omega}
\left(
\frac{\partial u}{\partial W}(x,W)\,\gamma
+
\mathcal{N}(u(x,W))
\right)
v_k(x)\,dx
\right]
\left[
\int_{\Omega}
\frac{\partial u}{\partial W}(x,W)\,v_k(x)\,dx
\right]
= 0.
\end{equation}
This expression indicates that the optimal direction $\gamma$ must balance the weak residuals and their sensitivities with respect to the parameters.

To obtain an explicit algebraic form, introduce for each test function $v_k$ the row vector $J_k\in\mathbb{R}^{P}$ and scalar $N_k$ defined by
\begin{equation}
J_{kj}
=
\int_{\Omega}
\frac{\partial u}{\partial W_j}(x,W)\,v_k(x)\,dx,
\qquad
N_k
=
\int_{\Omega}
\mathcal{N}(u(x,W))\,v_k(x)\,dx,
\qquad
1\le k\le K,\;1\le j\le P.
\end{equation}
The vector $J_k$ collects the weak sensitivities of the network output with respect to each parameter $W_j$ under the test function $v_k$, while $N_k$ captures the weak projection of the PDE residual under the test function $v_k(x)$.  
With these definitions, the optimality condition can be rewritten in compact index notation as
\begin{equation}
\sum_{k=1}^{K}
\Bigg[
\underbrace{
\left(
\int_{\Omega}
\frac{\partial u}{\partial W}(x,W)\,v_k(x)\,dx
\right)^{\!T}
}_{J_k^{T}}
\left(
\underbrace{
\left(
\int_{\Omega}
\frac{\partial u}{\partial W}(x,W)\,v_k(x)\,dx
\right)\gamma
}_{J_k\gamma}
+
\underbrace{
\int_{\Omega}
\mathcal{N}(u(x,W))\,v_k(x)\,dx
}_{N_k}
\right)
\Bigg]
= 0.
\end{equation}
The above representation clarifies that the linearization of the objective around $\gamma$ involves only inner products between the sensitivity vectors and the projected residuals.

In practical computation, the spatial integrals defining $J_{kj}$ and $N_k$ are evaluated numerically.  
Using Gauss--Legendre quadrature on $\Omega$ with nodes $\{x_i\}_{i=1}^{N_x}$ and weights $\{w_i\}_{i=1}^{N_x}$ leads to the approximations
\begin{equation}
J_{kj}
=
\int_{\Omega}
\frac{\partial u}{\partial W_j}(x,W)\,v_k(x)\,dx
\;\approx\;
\sum_{i=1}^{N_x}
w_i\,
\frac{\partial u}{\partial W_j}(x_i,W)\,
v_k(x_i),
\end{equation}
\begin{equation}
N_k
=
\int_{\Omega}
\mathcal{N}(u(x,W))\,v_k(x)\,dx
\;\approx\;
\sum_{i=1}^{N_x}
w_i\,
\mathcal{N}(u(x_i,W))\,
v_k(x_i).
\end{equation}
These formulas express each entry of $J_k$ and $N_k$ as a weighted sum over collocation points and provide a discrete representation compatible with automatic differentiation for $\partial u/\partial W_j$.

Stacking the rows $J_k$ into a matrix
\begin{equation}
J
=
\begin{pmatrix}
J_1^T\\
\vdots\\
J_K^T
\end{pmatrix}
\in\mathbb{R}^{K\times P},
\qquad
N
=
\begin{pmatrix}
N_1\\
\vdots\\
N_K
\end{pmatrix}
\in\mathbb{R}^{K},
\end{equation}
the optimality condition becomes the normal equation
\begin{equation}
J^{T}J\,\gamma_{\text{opt}} = -J^{T}N.
\label{eq:equation_form}
\end{equation}
Here, $K$ denotes the number of test functions and $P$ denotes the number of trainable parameters in the network.
The matrix $J^{T}J\in\mathbb{R}^{P\times P}$ and vector $J^{T}N\in\mathbb{R}^{P}$ define a linear system for the update direction $\gamma_{\text{opt}}$.  
Solving this system yields the parameter evolution direction that minimizes the weak-form residual in the least-squares sense for the chosen set of test functions and quadrature nodes.  
The number of test functions $K$ is selected to balance expressive power and
numerical stability.
The equation form $J^T J \gamma = -J^T N$ in Eq.~\eqref{eq:equation_form} is adopted for consistency with
existing evolutionary-network formulations and for notational clarity.
Although solving normal equations may amplify conditioning issues in general,
our numerical results indicate that the weak-form projection substantially
improves conditioning compared to strong-form enforcement.
Alternative solvers based on QR or SVD factorizations, as well as Tikhonov
regularization, are fully compatible with the proposed framework and can be
employed when higher numerical robustness is required.

\subsubsection{Computational complexity}

Table~\ref{tab:complexity_strong_vs_weak} compares the dominant computational
costs of strong-form and weak-form evolutionary approaches over $M$ time steps.
The analysis focuses on the per-step evaluation of the residual $N$ and sensitivity matrix $J$, as well as on the solution of the parameter-update least-squares problem, and reports their leading-order asymptotic scaling.

In the strong-form approach, the least-squares problem is posed over
$N_{\hat{u}}$ collocation points, yielding a Jacobian
$J \in \mathbb{R}^{N_{\hat{u}}\times P}$.
At each time step, the PDE residual is evaluated pointwise, requiring
higher-order spatial derivatives and incurring a cost
$\mathcal{O}(N_{\hat{u}}\,C_{\mathrm{PDE}}^{\mathrm{SF}})$.
Automatic differentiation is applied at all collocation points to assemble
the $J$, with cost
$\mathcal{O}(N_{\hat{u}}\,P\,C_{\mathrm{AD}})$.
Assuming a dense least-squares solve, the per-step solution cost scales as
$\mathcal{O}(N_{\hat{u}}P^2)$.
The cumulative cost over $M$ time steps is therefore
\begin{equation}
    \mathcal{O}\!\left(
M\left(
N_{\hat{u}}P^2
+ N_{\hat{u}}P C_{\mathrm{AD}}
+ N_{\hat{u}} C_{\mathrm{PDE}}^{\mathrm{SF}}
\right)
\right),
\end{equation}
indicating that both the system dimension and residual evaluation cost grow
with the number of collocation points.

In the weak-form approach, the residual is projected onto a fixed set of $K$
test functions, resulting in a parameter-update system
$J \in \mathbb{R}^{K\times P}$ whose size is independent of the data resolution.
The PDE residual is evaluated at $N_x$ quadrature points, with reduced
differential order due to integration by parts, leading to a cost
$\mathcal{O}(N_x\,C_{\mathrm{PDE}}^{\mathrm{WF}})$.
Automatic differentiation at quadrature points yields a Jacobian assembly cost
of $\mathcal{O}(N_x\,P\,C_{\mathrm{AD}})$.
Projection onto the test space introduces additional integration costs:
$\mathcal{O}(K N_x)$ for assembling the weak residual and
$\mathcal{O}(K N_x P)$ for constructing the projected Jacobian.
The per-step least-squares system formation scales as $\mathcal{O}(K P^2)$,
and the total cost over $M$ time steps becomes
\begin{equation}
\mathcal{O}\!\left(
M\left(
K P^2
+ N_x P C_{\mathrm{AD}}
+ N_x C_{\mathrm{PDE}}^{\mathrm{WF}}
+ K N_x P
+ K N_x
\right)
\right).
\end{equation}
Typically, the number of strong-form collocation points satisfies
$N_{\hat{u}} \gg K$, whereas in the weak formulation the projection is
evaluated using a similar number of quadrature points $N_x$ as the
collocation points $N_{\hat{u}}$, while the dimension of the resulting
update system remains governed by the number of test functions~$K$.
As a result, the weak-form evolutionary solver exhibits more favorable scaling,
with additional data primarily improving projection accuracy rather than
increasing the cost of the parameter-update system.
We emphasize that the computational advantage of the proposed weak-form
evolutionary solver lies in decoupling the parameter-update system size from
the number of quadrature points.
While increasing $N_x$ improves the accuracy of weak residual evaluation,
it does not alter the dimension of the least-squares system, which remains
fixed by the number of test functions $K$.

\begin{table*}[t]
\centering
\small
\setlength{\tabcolsep}{6pt}
\renewcommand{\arraystretch}{1.2}
\caption{Computational complexity comparison of strong-form and weak-form evolutionary solvers over $M$ time steps.
Here $J=\partial \hat u/\partial W$ and $N=\mathcal N(\hat u)$ denote the parameter Jacobian and PDE residual.
$N_{\hat u}$ and $N_x$ are the numbers of collocation and quadrature points, $K$ the number of test functions,
and $P$ the number of trainable parameters.
$C_{\mathrm{AD}}$ denotes parameter automatic differentiation cost, while $C_{\mathrm{PDE}}$ denotes spatial differentiation cost for residual evaluation.
Due to integration by parts, $C_{\mathrm{PDE}}^{\mathrm{WF}}$ is lower than $C_{\mathrm{PDE}}^{\mathrm{SF}}$.}
\label{tab:complexity_strong_vs_weak}

\begin{tabular}{@{}p{0.26\textwidth} p{0.34\textwidth} p{0.34\textwidth}@{}}
\toprule
 & Strong form & Weak form \\
\midrule
System to solve
&
$J^{T\,(P\times N_{\hat u})}
J^{(N_{\hat u}\times P)}\gamma_{\text{opt}}
=
- J^{T\,(P\times N_{\hat u})}
N^{(N_{\hat u}\times 1)}$
&
$J^{T\,(P\times K)}
J^{(K\times P)}\gamma_{\text{opt}}
=
- J^{T\,(P\times K)}
N^{(K\times 1)}$
\\


Data--system coupling
&
\textbf{Coupled}: system size scales with $N_{\hat{u}}$
&
\textbf{Decoupled}: system size fixed by $K$ \\

$J$ evaluation  
&
$\mathcal{O}(N_{\hat{u}}\,P\,C_{\mathrm{AD}})$
&
$\mathcal{O}(N_x\,P\,C_{\mathrm{AD}} + K N_x P)$ \\

$N$ evaluation
&
$\mathcal{O}(N_{\hat{u}}\,C_{\mathrm{PDE}}^{\mathrm{SF}})$
&
$\mathcal{O}(N_x\,C_{\mathrm{PDE}}^{\mathrm{WF}} + K N_x)$ \\


Least-squares system formation
&
$\mathcal{O}(N_{\hat{u}}P^2)$
&
$\mathcal{O}(K P^2)$ \\

\midrule
Total cumulative cost
&
$\mathcal{O}\!\left(
M\left(
N_{\hat{u}}P^2
+ N_{\hat{u}}P C_{\mathrm{AD}}
+ N_{\hat{u}} C_{\mathrm{PDE}}^{\mathrm{SF}}
\right)
\right)$
&
\makecell{$\mathcal{O}\!\left(
M\left(
K P^2
+ N_x P C_{\mathrm{AD}}
+ N_x C_{\mathrm{PDE}}^{\mathrm{WF}}
\right.\right.$\\
$\left.\left.
+ K N_x P
+ K N_x
\right)
\right)$} \\

\bottomrule
\end{tabular}
\end{table*}

\section{Numerical Experiments}
\label{Sec:numerical_experiments}

In this section, three approaches are compared:  
(i) EvoKAN-WF (Evolutionary KAN with weak-form, Sec.~\ref{sec:VEKAN}),  
(ii) EvoKAN-SF (Evolutionary KAN with strong-form, Sec.~\ref{sec:EvoKAN_RBFs}),  
(iii) a standard PINN-SF~\cite{raissi2019pinn} (PINN with strong-form).  
EvoKAN-WF adopts a weak formulation in which the residual is projected onto a
finite set of test functions, using Gaussian \ac{RBFs} together with the
boundary-constrained KAN described in Sec.~\ref{sec:BC}.  
EvoKAN-SF employs the same Gaussian \ac{RBFs} architecture and boundary
construction but enforces the governing equation in the strong form at
collocation points.  
The PINN-SF baseline is trained over the entire spatio-temporal domain and
applies boundary conditions through soft penalty terms without an
evolutionary update in time.
Regarding computational resources, an \texttt{NVIDIA GeForce RTX~4090} GPU was used for initial condition training, while parameter evolution ran on an \texttt{AMD Threadripper PRO~5955WX} with 31 cores.

\subsection{1D Allen-Cahn equation}

The one-dimensional Allen--Cahn equation considered in this numerical experiment is
\begin{equation}
\label{eq:AC_simple_eps}
\frac{\partial u}{\partial t}
=
\frac{\partial^{2}u}{\partial x^{2}}
-
\frac{1}{\epsilon^{2}}\,u\,(u^{2}-1),
\qquad x\in[-1,1],\; t\ge 0,
\end{equation}
where the parameter $\epsilon=0.002$ controls the thickness of the diffuse interface.  
Smaller values of $\epsilon$ produce sharper transition layers, resulting in a stiff nonlinear reaction term and rapid changes in the solution profile.

The initial condition is given by
\begin{equation}
u(x,0)=0.08\sin(\pi x),
\end{equation}
and homogeneous Dirichlet boundary conditions are imposed:
\begin{equation}
u(-1,t)=0,\qquad u(1,t)=0.
\end{equation}

To derive the weak formulation, the PDE is multiplied by a test function $v_k$ and integrated over the spatial domain:
\begin{equation}
\int_{\Omega}
\frac{\partial u}{\partial t}\,v_k\,dx
=
\int_{\Omega}
\frac{\partial^{2}u}{\partial x^{2}}\,v_k\,dx
-
\int_{\Omega}
\frac{1}{\epsilon^{2}}\,u\,(u^{2}-1)\,v_k\,dx.
\end{equation}

The diffusion term is treated by integration by parts:
\begin{equation}
\int_{\Omega}
\frac{\partial^{2}u}{\partial x^{2}}\,v_k\,dx
=
-\int_{\Omega}
\frac{\partial u}{\partial x}\,
\frac{\partial v_k}{\partial x}\,dx
+
\left[
\frac{\partial u}{\partial x}\,v_k
\right]_{x=-1}^{x=1}.
\end{equation}

Because the test functions satisfy $v_k(\pm 1)=0$, the boundary contribution
vanishes, and the weak form of the Allen--Cahn equation becomes
\begin{equation}
\int_{\Omega}
\frac{\partial u}{\partial t}\,v_k\,dx
=
-\int_{\Omega}
\frac{\partial u}{\partial x}\,
\frac{\partial v_k}{\partial x}\,dx
-
\int_{\Omega}
\frac{1}{\epsilon^{2}}\,u\,(u^{2}-1)\,v_k\,dx,
\qquad k=1,\ldots,K.
\end{equation}
In the numerical experiments, the test functions $v_k$ are chosen as
sinusoidal functions on $[-1,1]$ of the form
$v_k(x) = \sin\big(k\pi(x+1)/2\big)$, $k=1,\ldots,K$, which satisfy
$v_k(-1)=v_k(1)=0$ by construction and provide increasing spatial frequency
as $k$ increases.
Unless otherwise specified, all weak-form integrals are evaluated using
Gauss--Legendre quadrature with $N_x$ nodes per spatial dimension.
We verified that moderate increases in $N_x$ lead to negligible changes in
solution accuracy, confirming that the dominant computational cost is governed
by $K$ rather than the quadrature resolution.

We summarize the training settings for EvoKAN-WF, EvoKAN-SF, and PINN-SF in Table~\ref{table:1D_AC_training_setting}.
EvoKAN-WF and EvoKAN-SF employ the same network architecture and are trained for the initial condition using the Adam optimizer,
followed by time marching via parameter evolution with a fixed time step.
In contrast, PINN-SF uses a deeper MLP with \texttt{tanh} activations and is trained using Adam/L-BFGS-B without explicit time evolution.

\sisetup{group-separator={,}, group-minimum-digits=4}
\begin{table} [hbt!]
\small
	\renewcommand{\arraystretch}{1.0}
	\begin{center} 
		\caption{Training configuration for the 1D Allen-Cahn equation (Eq.~\eqref{eq:AC_simple_eps}).}
		\begin{tabular}{l c c c}
			\hline
			{\, \, \, } & \makecell[c]{EvoKAN-WF} & \makecell[c]{EvoKAN-SF} & {PINN-SF} \\
			\hline
			{Hidden layers} & {[3, 3, 3, 3]} & \makecell[c]{[3, 3, 3, 3]} & {[15, 15, 15]}\\
            {Activation functions} & \makecell[c]{Gaussian RBFs/SiLU} & \makecell[c]{Gaussian RBFs/SiLU} & {Tanh} \\
            \makecell[l]{Grid points number\\ of activation functions} & {4} & \makecell[c]{4}  & {-}\\
			\makecell[l]{Number of \\ trainable parameters} & {162} & {195} & \makecell[c]{526}\\
            {Optimizer} & \makecell[c]{Adam} & \makecell[c]{Adam} & \makecell[c]{Adam/L-BFGS-B}\\
            {Timestep} & \makecell[c]{1e-07} & \makecell[c]{1e-07} & \makecell[c]{-}\\
			\hline
		\end{tabular}
		\label{table:1D_AC_training_setting}
	\end{center}
\end{table}

\begin{figure}[h!]
    \centering
    \begin{subfigure}[b]{1\linewidth}
        \centering
        \includegraphics[width=\linewidth]{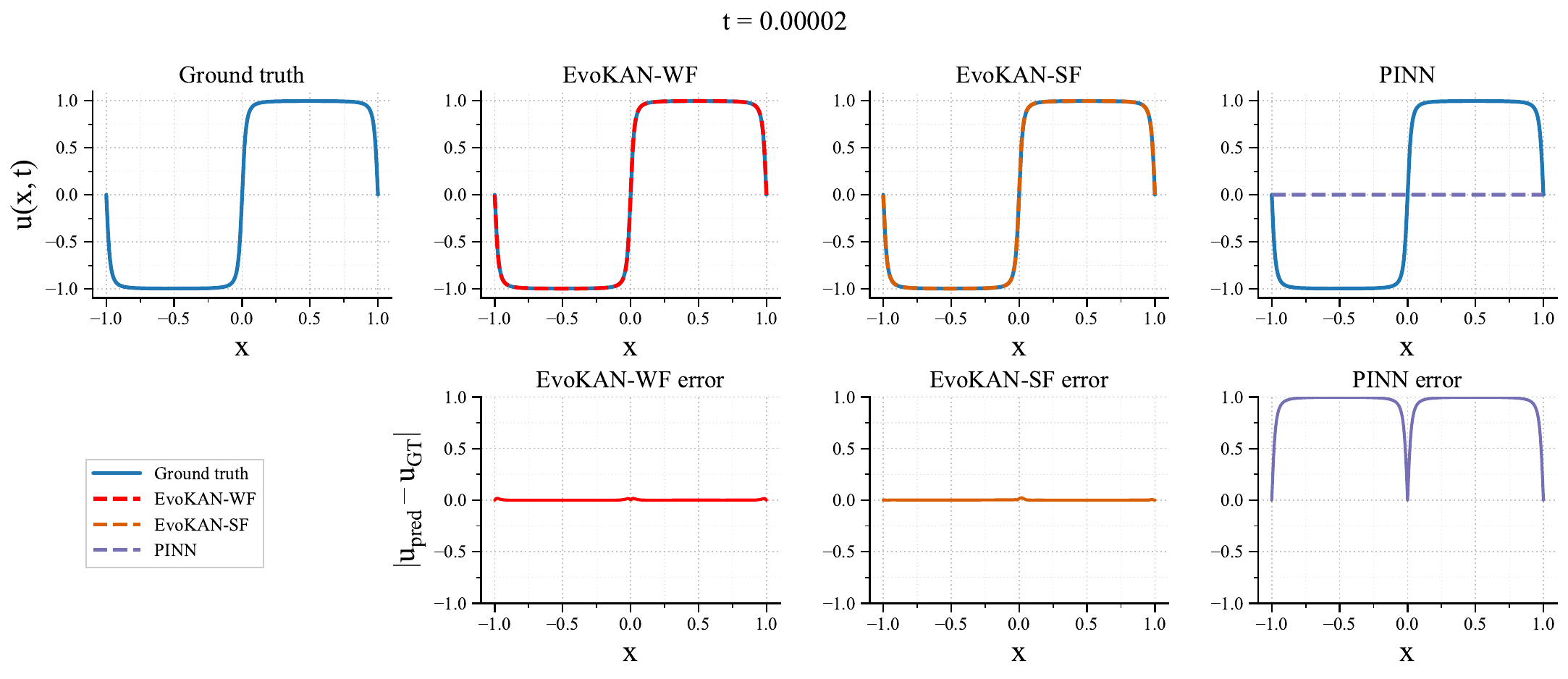}
    \end{subfigure} 
\caption{
Comparison of predicted solutions for the one-dimensional Allen--Cahn equation~(Eq.~\eqref{eq:AC_simple_eps}) at $t = 2 \times 10^{-5}$,
trained with 100 data points.
Both EvoKAN--WF and EvoKAN--SF closely match the ground-truth solution at this time,
whereas the vanilla PINN does not reproduce the correct profile.
}
\label{fig:1D_AC_solution_1}
\end{figure}

\begin{figure}[h!]
    \centering
    \begin{subfigure}[b]{1\linewidth}
        \centering
        \includegraphics[width=\linewidth]{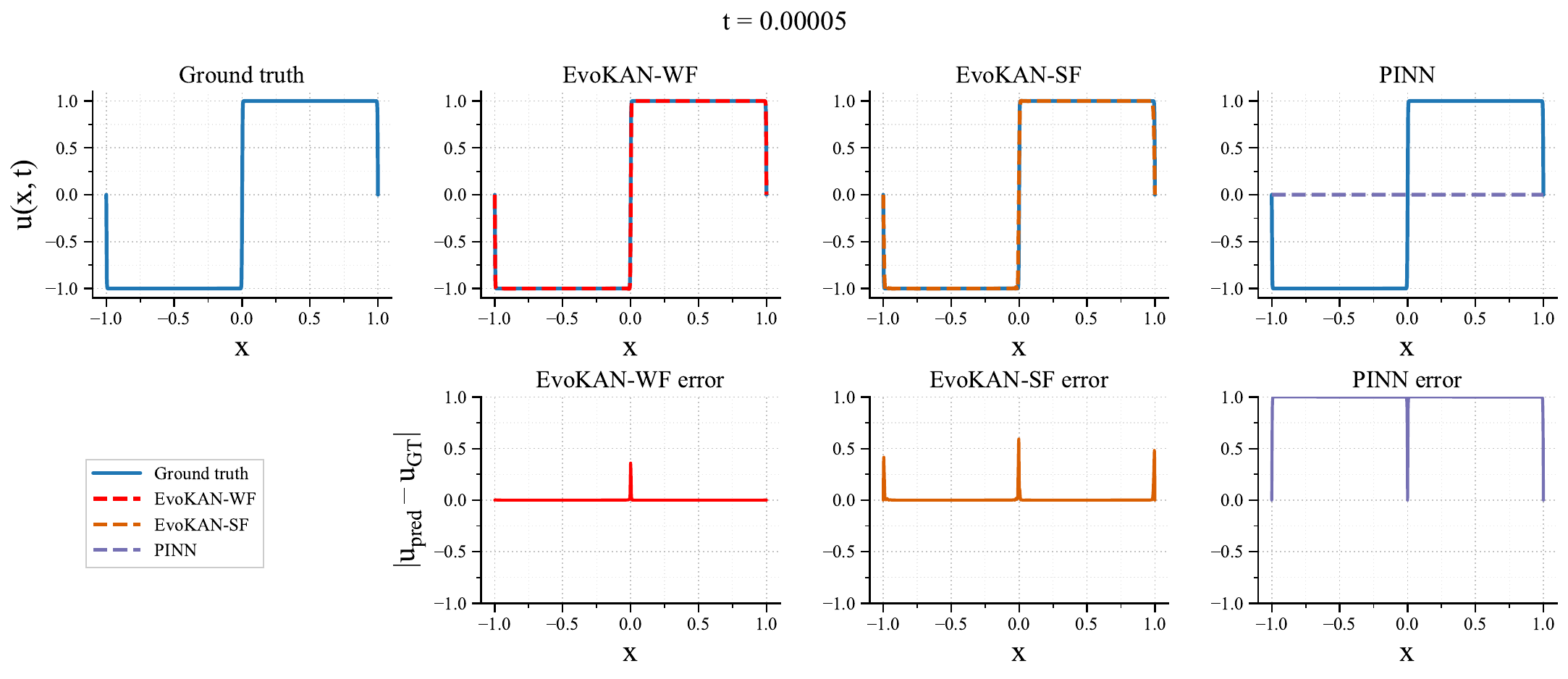}
    \end{subfigure} 
\caption{
Comparison of predicted solutions for the one-dimensional Allen--Cahn equation~(Eq.~\eqref{eq:AC_simple_eps}) at $t = 5 \times 10^{-5}$,
trained with 100 data points.
As the solution profile becomes steeper, EvoKAN--WF maintains good accuracy,
whereas EvoKAN--SF shows small deviations near the center and boundaries,
and the vanilla PINN fails to recover the correct solution.
}
\label{fig:1D_AC_solution_2}
\end{figure}

To compare the predictive behavior of the different solvers,
Figs.~\ref{fig:1D_AC_solution_1} and~\ref{fig:1D_AC_solution_2}
present the numerical solutions of the one-dimensional Allen--Cahn equation
at two representative time instances.
At the earlier time ($t = 2 \times 10^{-5}$, Fig.~\ref{fig:1D_AC_solution_1}),
both EvoKAN--WF and EvoKAN--SF closely match the ground-truth solution,
while the vanilla PINN does not accurately reproduce the profile.
At the later time ($t = 5 \times 10^{-5}$, Fig.~\ref{fig:1D_AC_solution_2}),
as the solution becomes steeper, EvoKAN--WF maintains good agreement with the reference solution,
whereas EvoKAN--SF exhibits localized deviations near the center and boundaries.
The vanilla PINN again fails to capture the correct solution behavior.
Overall, these results show that the weak-form evolutionary formulation
provides stable and accurate predictions across the tested time instances,
while the strong-form variant is slightly more sensitive to evolving solution gradients.

\begin{figure}[h!]
    \centering
    \begin{subfigure}[b]{0.32\linewidth}
        \centering
        \includegraphics[width=\linewidth]{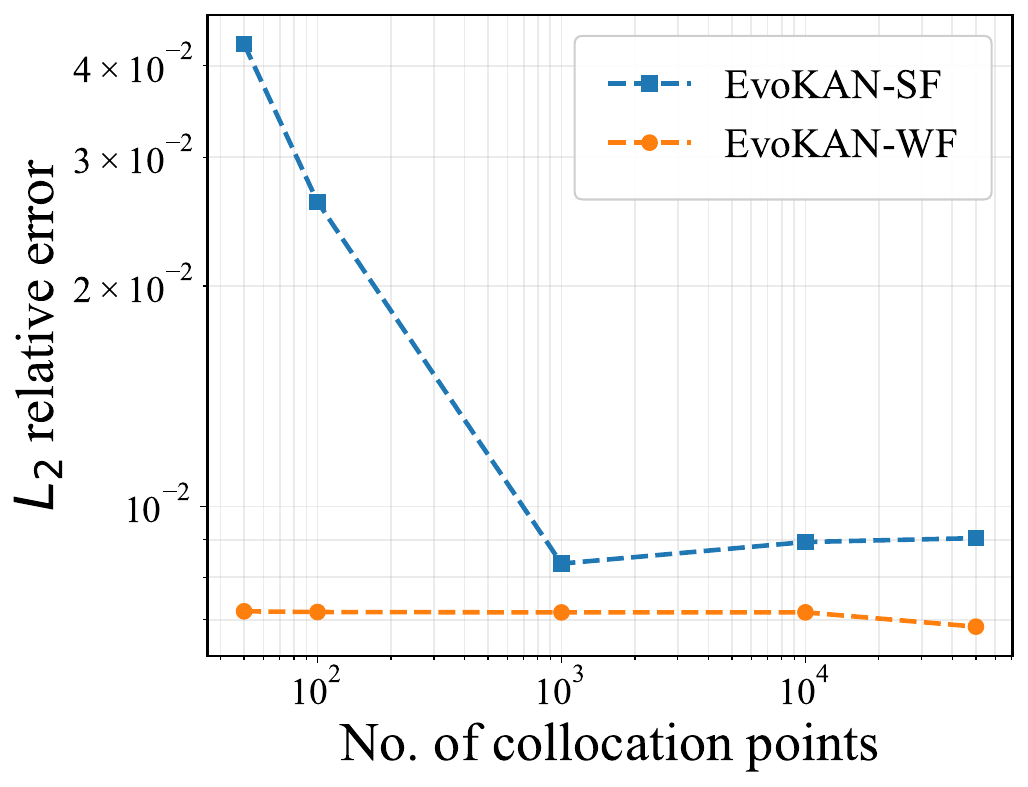}
        \subcaption{$L_2$ relative error}
    \end{subfigure} 
        \begin{subfigure}[b]{0.32\linewidth}
        \centering
        \includegraphics[width=\linewidth]{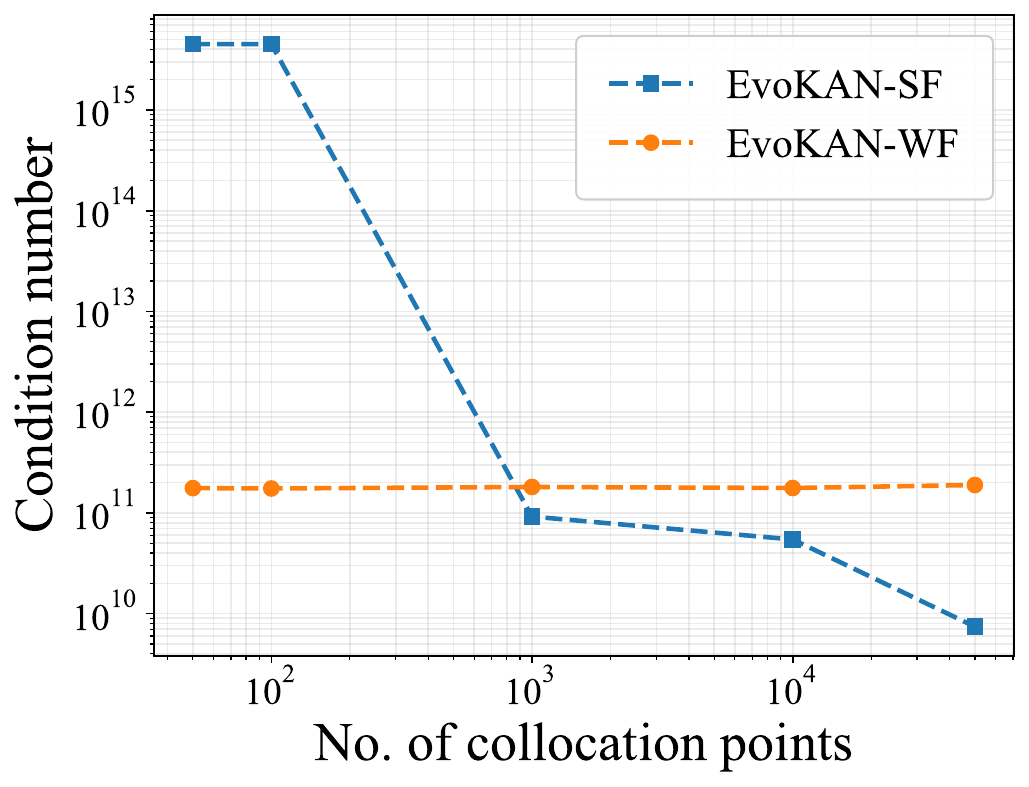}
        \subcaption{Condition number}
    \end{subfigure} 
        \begin{subfigure}[b]{0.32\linewidth}
        \centering
        \includegraphics[width=\linewidth]{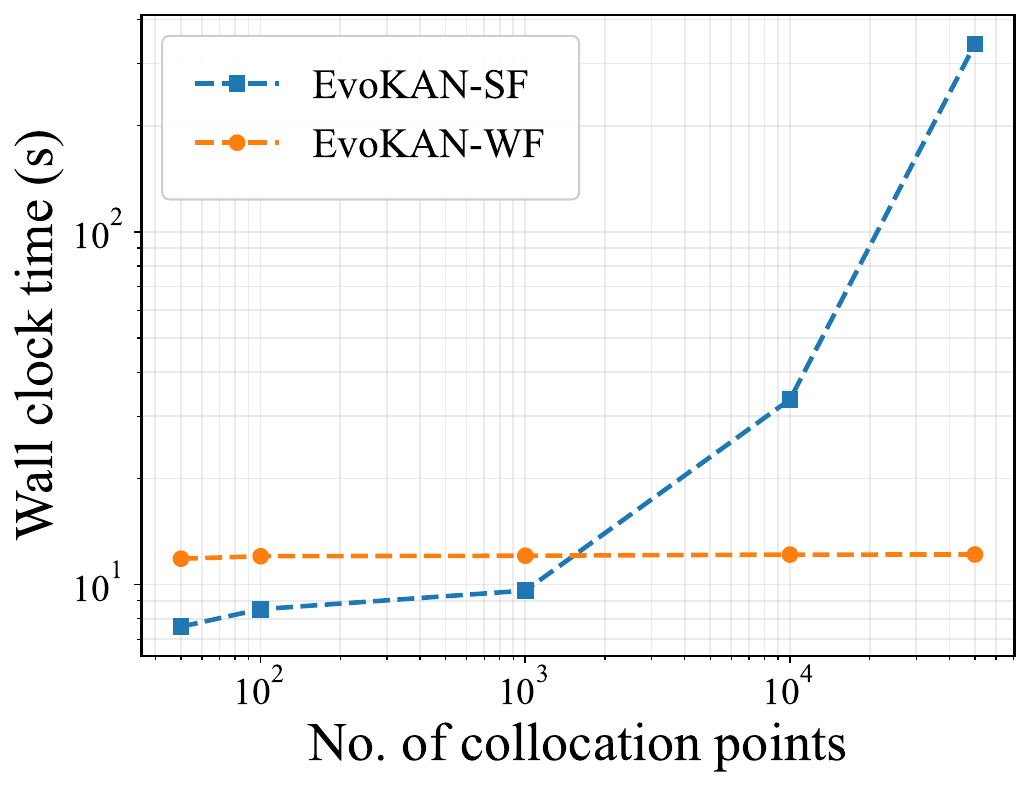}
        \subcaption{Computation time}
    \end{subfigure} 
        \caption{
Strong-form (EvoKAN-SF) versus weak-form (EvoKAN-WF) evolutionary KAN solvers
for the 1D Allen--Cahn equation (Eq.~\eqref{eq:AC_simple_eps}) 
with respect to the number of collocation points:
(a) $L_2$ relative error,
(b) condition number,
and (c) wall-clock time.
The weak-form approach achieves lower errors and substantially better conditioning,
while avoiding the rapid growth in computational cost observed in the strong-form solver.
}
\label{fig:1D_AC_metric}
\end{figure}

To examine the effect of the number of collocation points on accuracy, conditioning, and computational cost,
Figure~\ref{fig:1D_AC_metric} together with Table~\ref{table:1D_AC_error_time}
summarizes the $L_2$ relative error, condition number, and wall clock time
for EvoKAN SF and EvoKAN WF applied to the one dimensional Allen--Cahn equation.
As the number of collocation points increases, EvoKAN SF shows a non monotonic error trend
together with a growth of the condition number, which indicates ill conditioning
of the parameter update system.
In contrast, EvoKAN WF maintains low errors and a condition number
that remains bounded across all tested resolutions, reflecting the decoupling
between the update system size and the number of collocation points
in the weak formulation.
The computational cost of EvoKAN SF increases with the number of collocation points,
whereas EvoKAN WF shows limited variation across all tested resolutions.

\begin{table}[t]
\footnotesize
\centering
\caption{
$L_2$ relative error and wall clock time for the one-dimensional Allen--Cahn equation (Eq.~\eqref{eq:AC_simple_eps})
with varying the number of collocation points.
Speedup is defined as $T_{\text{SF}}/T_{\text{WF}}$.
Entries marked with $\dagger$ correspond to EvoKAN-SF solutions whose
$L_2$ relative error remains above $\mathcal{O}(10^{-2})$,
indicating insufficient accuracy.
}
\label{tab:error_time_allen_cahn}
\renewcommand{\arraystretch}{1.25}
\begin{tabular}{c cc cc c}
\toprule
& \multicolumn{2}{c}{$L_2$ relative error}
& \multicolumn{3}{c}{Wall clock time} \\
\cmidrule(lr){2-3} \cmidrule(lr){4-6}
\makecell{Number of \\ collocation points}
& EvoKAN-SF
& EvoKAN-WF
& EvoKAN-SF [s]
& EvoKAN-WF [s]
& Speedup \\
\midrule
64
& 4.2814e-02
& \textbf{7.1881e-03}
& \textbf{7.4396}$^{\dagger}$
& 11.8412
& \textbf{0.63} \\
100
& 2.6065e-02
& \textbf{7.1722e-03}
& \textbf{8.5188}$^{\dagger}$
& 12.0379
& \textbf{0.71} \\
1{,}000
& 8.3525e-03
& \textbf{7.1633e-03}
& 16.1555
& \textbf{12.0761}
& \textbf{1.34} \\
10{,}000
& 8.9389e-03
& \textbf{7.1644e-03}
& 33.5028
& \textbf{12.1654}
& \textbf{2.76} \\
50{,}000
& 9.0494e-03
& \textbf{6.8493e-03}
& 340.6359
& \textbf{12.1727}
& \textbf{27.99} \\
\bottomrule
\end{tabular}
\label{table:1D_AC_error_time}
\end{table}

\subsection{Dirichlet Boundary Condition: 2D Burgers' Equation}

The two-dimensional vector Burgers' equation on a square domain is written in strong form as
\begin{equation}
\label{eq:burgers_strong}
\begin{cases}
\displaystyle
\frac{\partial u}{\partial t}
=
\nu\left(
\frac{\partial^2 u}{\partial x^2}
+
\frac{\partial^2 u}{\partial y^2}
\right)
-
\left(
u\,\frac{\partial u}{\partial x}
+
v\,\frac{\partial u}{\partial y}
\right), \\[6pt]
\displaystyle
\frac{\partial v}{\partial t}
=
\nu\left(
\frac{\partial^2 v}{\partial x^2}
+
\frac{\partial^2 v}{\partial y^2}
\right)
-
\left(
u\,\frac{\partial v}{\partial x}
+
v\,\frac{\partial v}{\partial y}
\right),
\end{cases}
\qquad (x,y)\in\Omega,\; t>0,
\end{equation}
where $u(x,y,t)$ and $v(x,y,t)$ denote the two velocity components and $\nu>0$ is a constant viscosity.
The computational domain is defined as
\[
\Omega = (-1,1)\times(-1,1).
\]

The domain is equipped with homogeneous Dirichlet boundary conditions for both components,
\begin{equation}
\label{eq:burgers_bc}
u(x,y,t)=0,
\qquad
v(x,y,t)=0,
\qquad (x,y)\in\partial\Omega,\; t>0.
\end{equation}

To derive the weak formulation, a family of test functions
$\{v_k\}_{k=1}^K$ is introduced.
Each $v_k$ is constructed from a Fourier sine basis with periods chosen so that
$v_k=0$ on $\partial\Omega$, thereby satisfying the homogeneous Dirichlet boundary
condition by construction.
Multiplying the first equation in~\eqref{eq:burgers_strong} by $v_k$ and integrating over $\Omega$ yields
\[
\int_{\Omega}
\frac{\partial u}{\partial t}\,v_k\,dx\,dy
=
\int_{\Omega}
\nu\left(
\frac{\partial^2 u}{\partial x^2}
+
\frac{\partial^2 u}{\partial y^2}
\right)
v_k\,dx\,dy
-
\int_{\Omega}
\left(
u\,\frac{\partial u}{\partial x}
+
v\,\frac{\partial u}{\partial y}
\right)
v_k\,dx\,dy.
\]

The diffusion term is integrated by parts once, while the nonlinear advection term is left in advective form.
Writing the Laplacian as $\Delta u = \nabla\cdot(\nabla u)$ and applying integration by parts gives
\[
\int_{\Omega}
\nu\,\Delta u\,v_k\,dx\,dy
=
-\int_{\Omega}
\nu\,\nabla u\cdot\nabla v_k\,dx\,dy
+
\int_{\partial\Omega}
\nu\,\frac{\partial u}{\partial n}\,v_k\,ds.
\]
The boundary integral vanishes because the test functions are constructed
from a Fourier sine basis whose period is chosen so that
$v_k = 0$ on $\partial\Omega$.
As a result, the trace of each test function is zero on the boundary,
and the boundary contribution arising from integration by parts is eliminated.
Substituting this expression back into the balance yields the weak form for the $u$-component,
\begin{equation}
\label{eq:burgers_weak_component_u}
\int_{\Omega}
\frac{\partial u}{\partial t}\,v_k\,dx\,dy
+
\int_{\Omega}
\nu\,\nabla u\cdot\nabla v_k\,dx\,dy
+
\int_{\Omega}
\left(
u\,\frac{\partial u}{\partial x}
+
v\,\frac{\partial u}{\partial y}
\right)
v_k\,dx\,dy
=0,
\qquad k=1,\ldots,K.
\end{equation}

Applying the same procedure to the second equation in~\eqref{eq:burgers_strong} gives
\[
\int_{\Omega}
\frac{\partial v}{\partial t}\,v_k\,dx\,dy
=
\int_{\Omega}
\nu\left(
\frac{\partial^2 v}{\partial x^2}
+
\frac{\partial^2 v}{\partial y^2}
\right)
v_k\,dx\,dy
-
\int_{\Omega}
\left(
u\,\frac{\partial v}{\partial x}
+
v\,\frac{\partial v}{\partial y}
\right)
v_k\,dx\,dy.
\]
Integrating the diffusion term by parts yields
\[
\int_{\Omega}
\nu\,\Delta v\,v_k\,dx\,dy
=
-\int_{\Omega}
\nu\,\nabla v\cdot\nabla v_k\,dx\,dy
+
\int_{\partial\Omega}
\nu\,\frac{\partial v}{\partial n}\,v_k\,ds,
\]
and the boundary contribution vanishes due to $v_k=0$ on $\partial\Omega$.
The resulting weak form for the $v$-component is
\begin{equation}
\label{eq:burgers_weak_component_v}
\int_{\Omega}
\frac{\partial v}{\partial t}\,v_k\,dx\,dy
+
\int_{\Omega}
\nu\,\nabla v\cdot\nabla v_k\,dx\,dy
+
\int_{\Omega}
\left(
u\,\frac{\partial v}{\partial x}
+
v\,\frac{\partial v}{\partial y}
\right)
v_k\,dx\,dy
=0,
\qquad k=1,\ldots,K.
\end{equation}
Equations~\eqref{eq:burgers_weak_component_u} and~\eqref{eq:burgers_weak_component_v}
together define the weak formulation of the two-dimensional Burgers' equation
under homogeneous Dirichlet boundary conditions
and form the basis for the numerical experiments.

\sisetup{group-separator={,}, group-minimum-digits=4}
\begin{table} [hbt!]
\footnotesize
	\renewcommand{\arraystretch}{1.0}
	\begin{center} 
		\caption{Training configuration for the 2D Burgers equation (Eq.~\eqref{eq:burgers_strong}).}
		\begin{tabular}{l c c c}
			\hline
			{\, \, \, } & \makecell[c]{EvoKAN-WF} & \makecell[c]{EvoKAN-SF} & {PINN-SF} \\
			\hline
			{Hidden layers} & {[16, 16, 16]} & \makecell[c]{[16, 16, 16]} & {[25, 25, 25]}\\
            {Activation functions} & \makecell[c]{Gaussian RBFs/SiLU} & \makecell[c]{Gaussian RBFs/SiLU} & {Tanh} \\
            \makecell[l]{Grid points number\\ of activation functions} & {5} & \makecell[c]{5}  & {-}\\
			\makecell[l]{Number of \\ trainable parameters} & {3392} & {3472} & \makecell[c]{3670}\\
            {Optimizer} & \makecell[c]{Adam} & \makecell[c]{Adam} & \makecell[c]{Adam/L-BFGS-B}\\
            {Timestep} & \makecell[c]{1e-03} & \makecell[c]{1e-03} & \makecell[c]{-}\\
			\hline
		\end{tabular}
		\label{table:burgers_training}
	\end{center}
\end{table}

The training settings for the two dimensional Burgers equation
are summarized in Table~\ref{table:burgers_training}.
EvoKAN-WF and EvoKAN-SF use the same network architecture
and are advanced in time using Adam with a fixed time step.
PINN-SF employs a wider MLP with tanh activations
and is trained using Adam and L-BFGS-B without time evolution.

\begin{figure}[h!]
    \centering
    
    \begin{subfigure}[b]{1\linewidth}
        \centering
        \includegraphics[width=\linewidth]{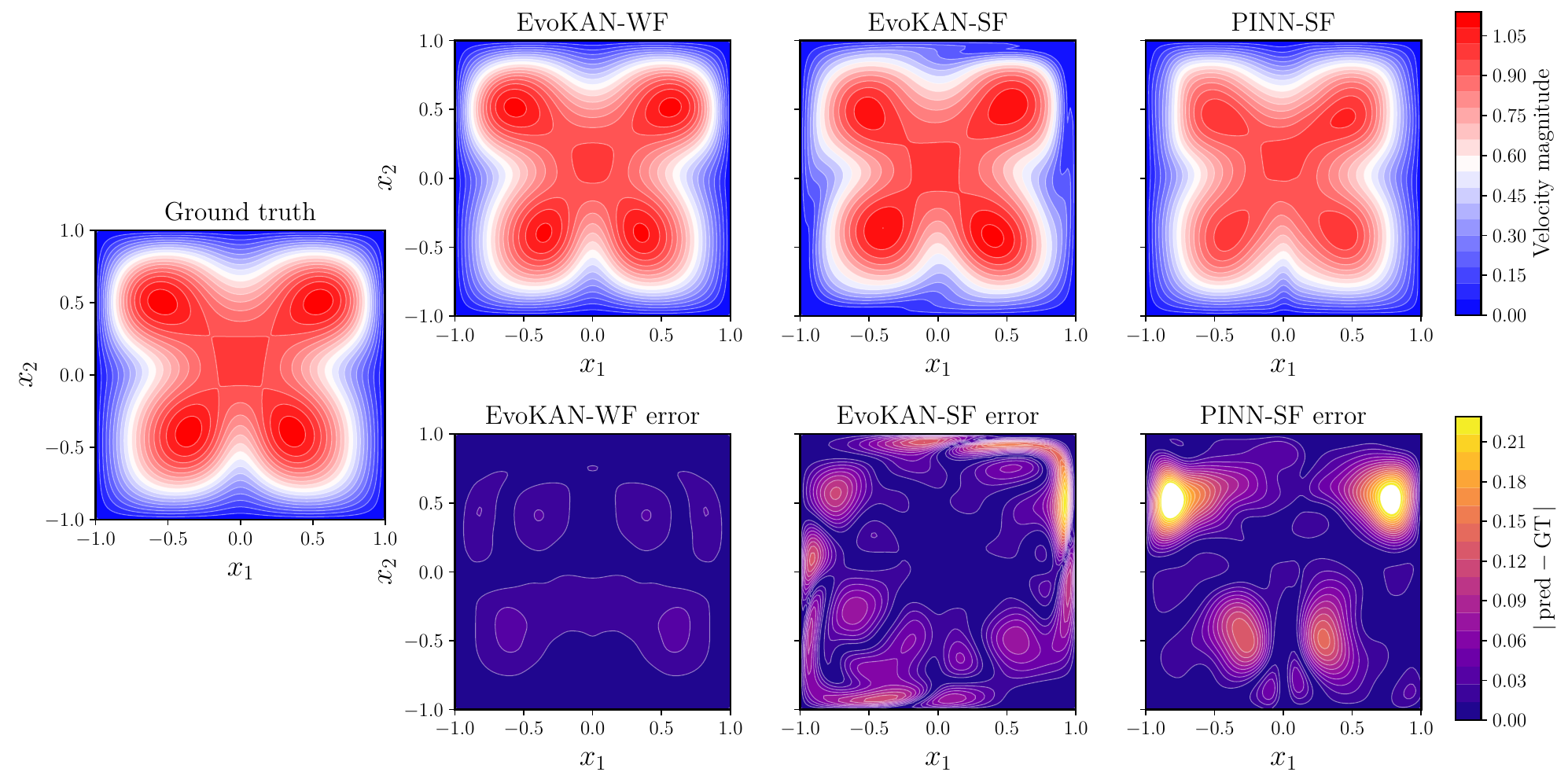}
    \end{subfigure}
\caption{
\textbf{Comparison of two-dimensional Burgers' equation (Eq.~\eqref{eq:burgers_strong}) solutions at $t = 0.1$.}
All models are trained using $900$ collocation points.
The top row shows the ground-truth FDM solution together with predictions from EvoKAN-WF, EvoKAN-SF, and PINN-SF.
The bottom row presents the corresponding absolute error fields with respect to the ground truth.
At this early time, EvoKAN-WF yields smaller errors than the other models
while providing comparable representations of the overall flow structure.
}
\label{fig:2D_Burgers_solution_01}
\end{figure}

\begin{figure}[h!]
    \centering
    
    \begin{subfigure}[b]{1\linewidth}
        \centering
        \includegraphics[width=\linewidth]{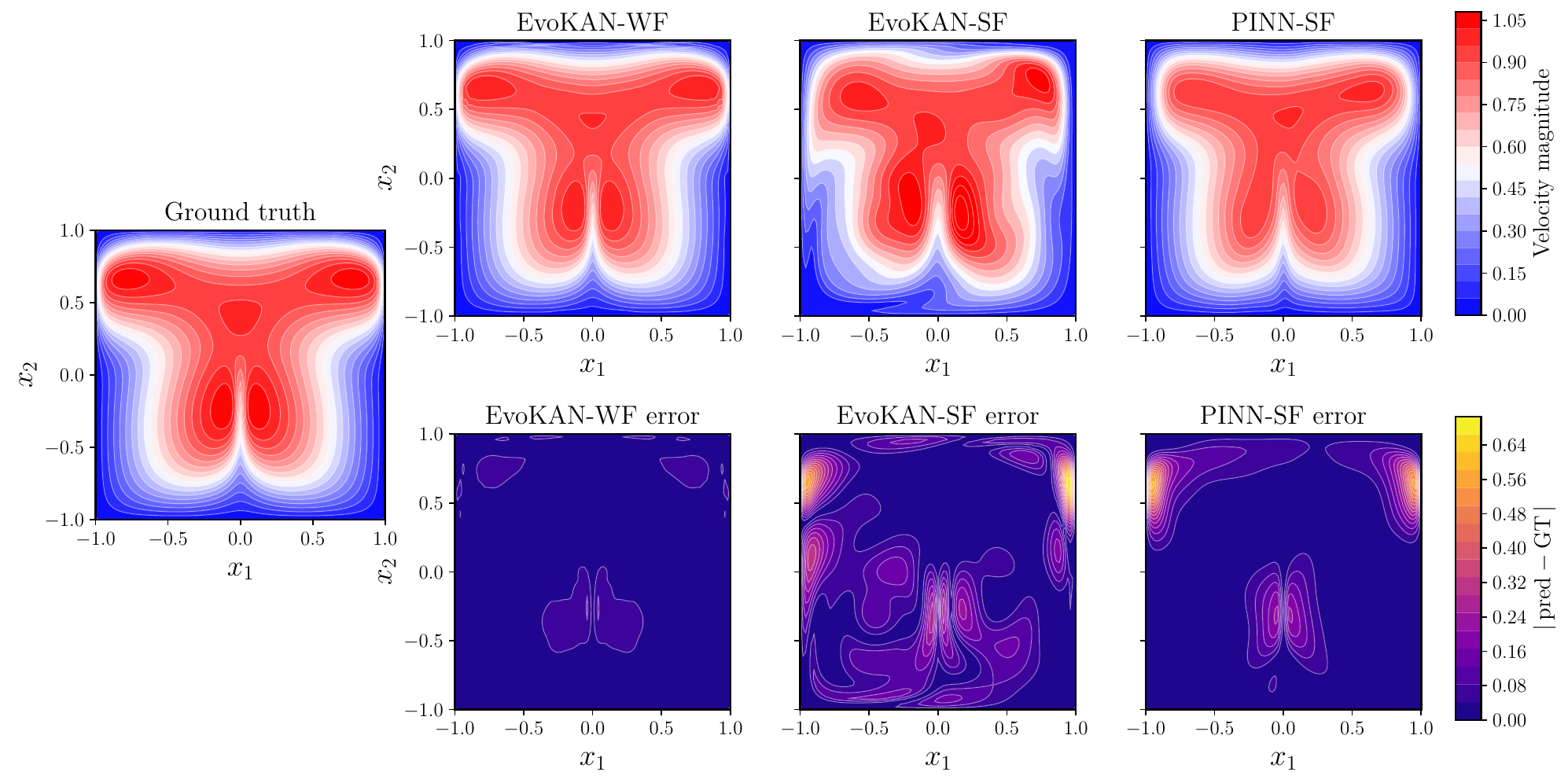}
    \end{subfigure}
\caption{
\textbf{Comparison of two-dimensional Burgers' equation (Eq.~\eqref{eq:burgers_strong}) solutions at $t = 0.4$.}
All models are trained using $900$ collocation points.
The top row shows the ground-truth FDM solution together with predictions from EvoKAN-WF, EvoKAN-SF, and PINN-SF.
The bottom row presents the corresponding absolute error fields with respect to the ground truth.
Despite the limited number of collocation samples, EvoKAN-WF provides more accurate predictions than the other models
and better captures the flow structure across the domain.
}
\label{fig:2D_Burgers_solution_04}
\end{figure}

To illustrate the effect of the weak formulation,
Fig.~\ref{fig:2D_Burgers_solution_01} and Fig.~\ref{fig:2D_Burgers_solution_04}
compare the predicted velocity magnitude fields
for the two dimensional Burgers equation at $t=0.1$ and $t=0.4$, respectively.
All models are trained using $900$ collocation points.
At both time instances, EvoKAN-WF reproduces the main flow structures observed
in the reference FDM solution.
The corresponding error fields remain spatially distributed
without pronounced localization.
EvoKAN-SF and PINN-SF predict the overall flow pattern,
while exhibiting larger error amplitudes
in regions associated with stronger velocity gradients.
At the later time $t=0.4$, differences between the methods become more apparent.
EvoKAN-WF continues to represent the dominant flow features across the domain,
whereas EvoKAN-SF and PINN-SF show increased discrepancies
in localized regions.
These results indicate that the weak-form formulation remains computationally efficient by maintaining predictive accuracy even with a relatively small number of collocation points, while preserving the global structure of the solution.

\begin{figure}[h!]
    \centering
    \begin{subfigure}[b]{0.32\linewidth}
        \centering
        \includegraphics[width=\linewidth]{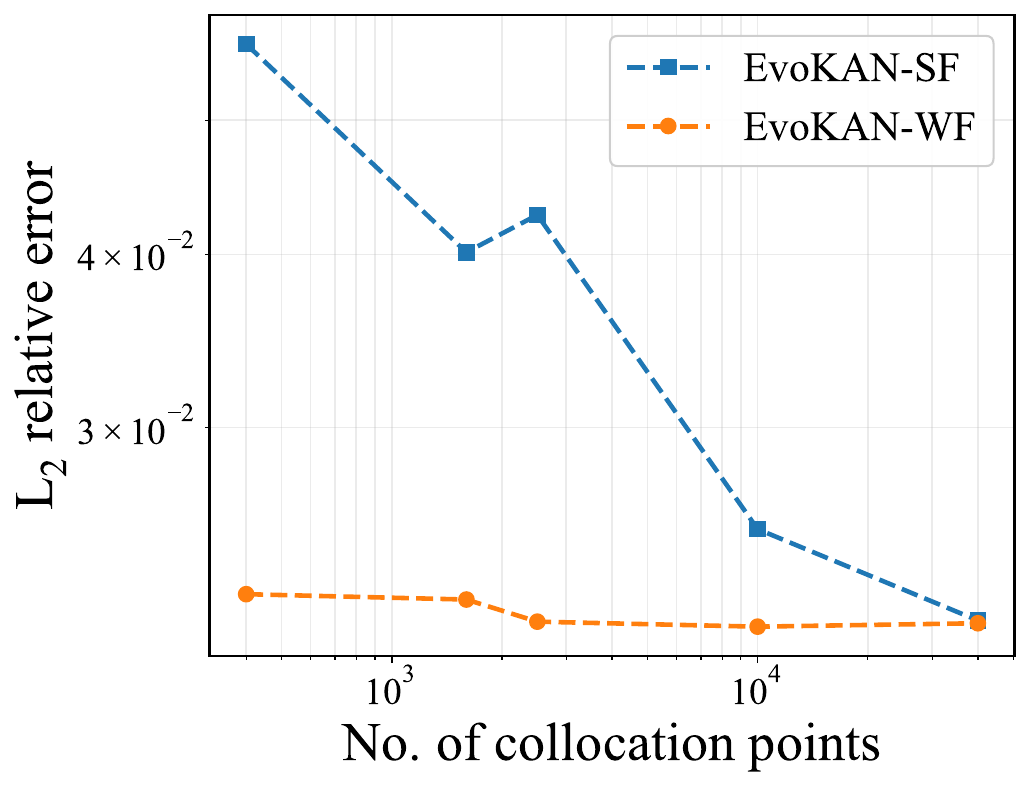}
        \subcaption{$L_2$ relative error}
    \end{subfigure} 
        \begin{subfigure}[b]{0.32\linewidth}
        \centering
        \includegraphics[width=\linewidth]{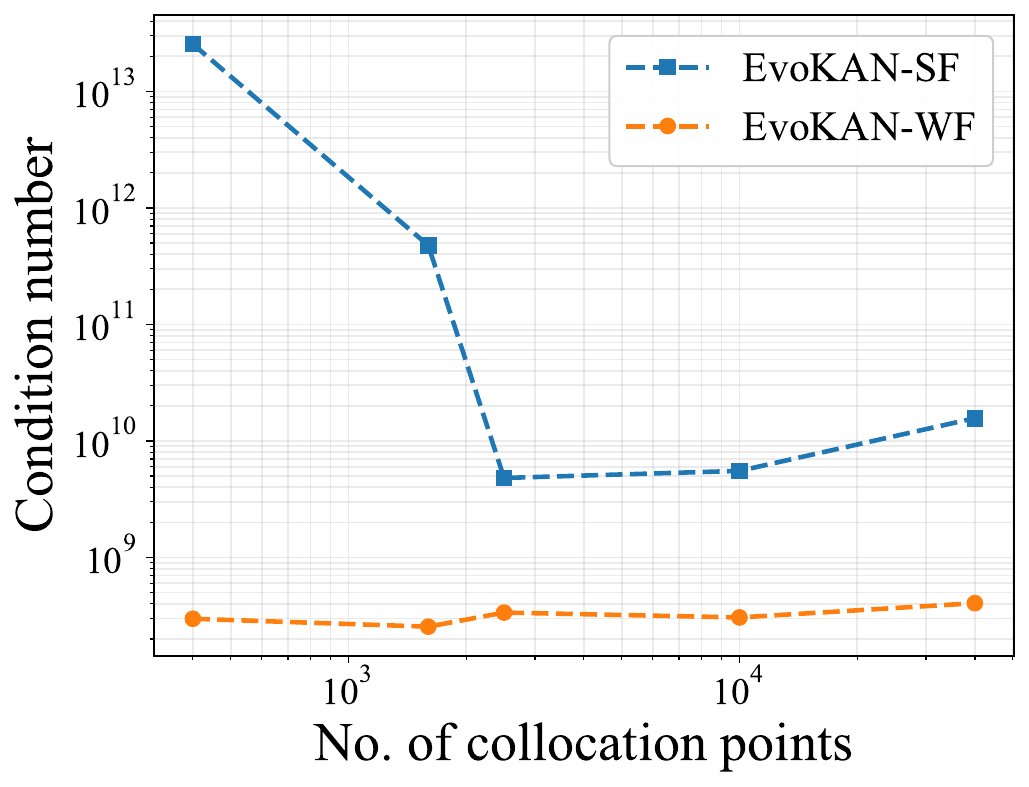}
        \subcaption{Condition number}
    \end{subfigure} 
        \begin{subfigure}[b]{0.32\linewidth}
        \centering
        \includegraphics[width=\linewidth]{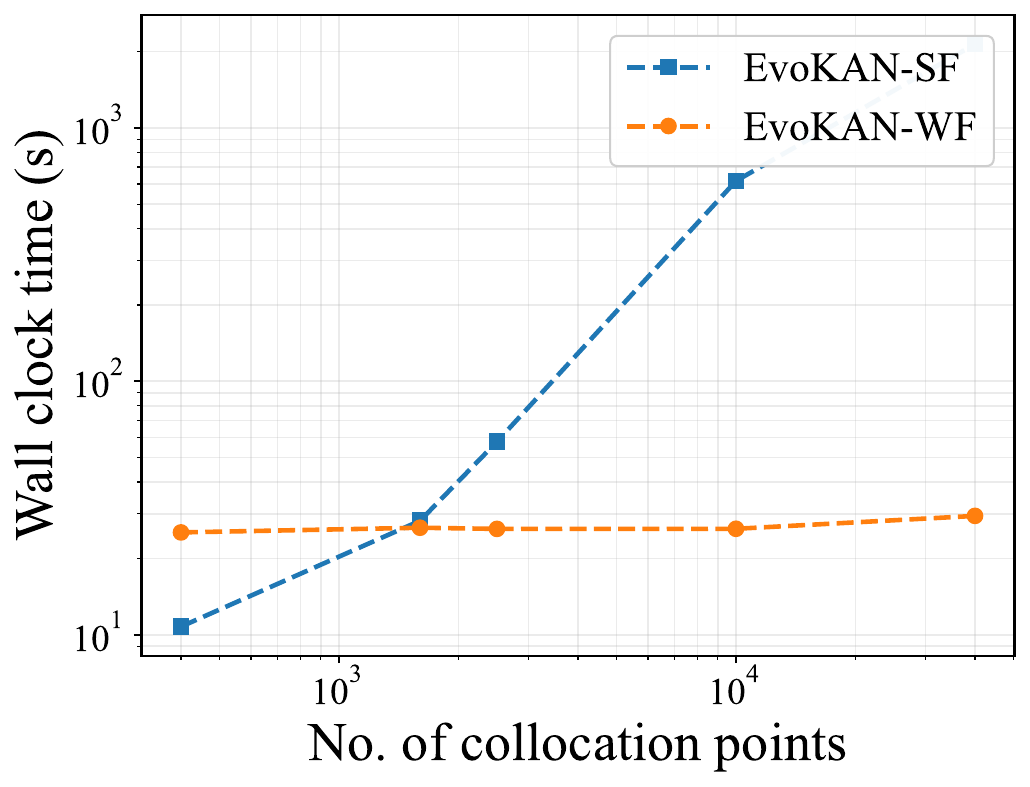}
        \subcaption{Computation time}
    \end{subfigure} 
        \caption{
Strong-form (EvoKAN-SF) versus weak-form (EvoKAN-WF) evolutionary KAN solvers
for the two-dimensional Burgers' equation (Eq.~\eqref{eq:burgers_strong})
with respect to the number of collocation points:
(a) $L_2$ relative error,
(b) condition number,
and (c) wall-clock time.
The weak-form approach achieves lower errors and improved numerical conditioning,
while avoiding the rapid growth in computational cost observed in the strong-form solver.
}
\label{fig:2D_Burgers_summary}
\end{figure}

\begin{table}[t]
\footnotesize
\centering
\caption{
$L_2$ relative error and wall clock time for the two-dimensional Burgers' equation (Eq.~\eqref{eq:burgers_strong})
with varying the number of collocation points.
Speedup is defined as $T_{\text{SF}}/T_{\text{WF}}$.
Entries marked with $\dagger$ correspond to EvoKAN-SF solutions whose
$L_2$ relative error exceeds $5\%$, indicating insufficient accuracy.
}
\label{tab:error_time_burgers}
\renewcommand{\arraystretch}{1.25}
\begin{tabular}{c cc cc c}
\toprule
& \multicolumn{2}{c}{$L_2$ relative error}
& \multicolumn{3}{c}{Wall clock time} \\
\cmidrule(lr){2-3} \cmidrule(lr){4-6}
\makecell{Number of \\ collocation points}
& EvoKAN-SF
& EvoKAN-WF
& EvoKAN-SF [s]
& EvoKAN-WF [s]
& Speedup \\
\midrule
900
& 5.67383e-02
& \textbf{2.27452e-02}
& \textbf{10.7565}$^{\dagger}$
& 25.3241
& \textbf{0.42} \\
1{,}600
& 4.01466e-02
& \textbf{2.25431e-02}
& 28.1469
& \textbf{26.4252}
& \textbf{1.07} \\
2{,}500
& 4.27184e-02
& \textbf{2.17294e-02}
& 57.7473
& \textbf{26.1492}
& \textbf{2.21} \\
10{,}000
& 2.53424e-02
& \textbf{2.15498e-02}
& 617.0453
& \textbf{26.1723}
& \textbf{23.58} \\
40{,}000
& 2.17728e-02
& \textbf{2.16729e-02}
& 2139.5068
& \textbf{29.4292}
& \textbf{72.69} \\
\bottomrule
\end{tabular}
\label{table:2D_Burgers}
\end{table}

The quantitative trends observed in Fig.~\ref{fig:2D_Burgers_summary}
are further supported by the numerical results reported in Table~\ref{table:2D_Burgers}.
For small sample sizes, EvoKAN-SF attains lower wall clock time
but fails to achieve sufficient accuracy, as indicated by the large $L_2$ errors.
As the number of collocation points increases,
EvoKAN-WF consistently maintains lower errors
while the computational cost remains nearly unchanged.
In contrast, the wall clock time of EvoKAN-SF grows rapidly,
leading to a substantial speedup in favor of EvoKAN-WF
for moderate to large sample sizes.
These results confirm the stable accuracy and favorable scaling behavior
of the weak form approach for the two dimensional Burgers' equation.

\subsection{2D Heat Equation with Nonlinear Forcing Term}

The two-dimensional heat equation with a nonlinear reaction term is written in strong form as
\begin{equation}
\label{eq:Heat_2D_text}
\frac{\partial u}{\partial t}(x,y,t)
=
\alpha \left(
\frac{\partial^{2}u}{\partial x^{2}}(x,y,t)
+
\frac{\partial^{2}u}{\partial y^{2}}(x,y,t)
\right)
+
u(t,x,y)\bigl(1-u(x,y,t)\bigr),
\qquad (x,y)\in\Omega,
\end{equation}
where the spatial domain is defined as $\Omega = (-1,1)^2$ and $\alpha>0$ denotes the diffusion coefficient.
In the present numerical experiment, the diffusion coefficient is fixed to $\alpha=0.1$ so that diffusion
and nonlinear reaction act on comparable spatial and temporal scales.

The initial condition prescribes the state at $t=0$ as
\begin{equation}
u(x,y,0) = \cos(\pi x)\cos(\pi y),
\qquad (x,y)\in\Omega,
\end{equation}
and homogeneous Neumann boundary conditions enforce zero normal flux on the boundary $\partial\Omega$,
\begin{equation}
\frac{\partial u}{\partial n}(x,y,t) = 0,
\qquad (x,y)\in\partial\Omega,\; t>0,
\end{equation}
where $\partial u/\partial n$ denotes the outward normal derivative.

To derive the weak formulation on $\Omega$, a finite set of test functions
$\{v_k(x,y)\}_{k=1}^{K}$ is introduced.
Multiplying the strong form~\eqref{eq:Heat_2D_text} by $v_k$ and integrating over $\Omega$ yields
\begin{equation}
\int_{\Omega}
u_t\,v_k\,d\Omega
=
\alpha\int_{\Omega}
\Delta u\,v_k\,d\Omega
+
\int_{\Omega}
u(1-u)\,v_k\,d\Omega.
\end{equation}
Using integration by parts for the diffusion term gives
\begin{equation}
\int_{\Omega}
\Delta u\,v_k\,d\Omega
=
\int_{\partial\Omega}
\frac{\partial u}{\partial n}\,v_k\,d\Gamma-\int_{\Omega}
\nabla u\cdot\nabla v_k\,d\Omega
.
\end{equation}
The boundary integral vanishes due to the homogeneous Neumann condition, and the weak form becomes
\begin{equation}
\label{eq:Heat_weak_2D_text}
\int_{\Omega}
u_t\,v_k\,d\Omega
+
\alpha\int_{\Omega}
\nabla u\cdot\nabla v_k\,d\Omega
-
\int_{\Omega}
u(1-u)\,v_k\,d\Omega
=0,
\qquad k=1,\ldots,K.
\end{equation}

\sisetup{group-separator={,}, group-minimum-digits=4}
\begin{table} [hbt!]
\footnotesize
	\renewcommand{\arraystretch}{1.0}
	\begin{center} 
		\caption{Training configuration for the 2D Heat equation with nonlinear forcing term (Eq.~\eqref{eq:Heat_2D_text}).}
		\begin{tabular}{l c c c}
			\hline
			{\, \, \, } & \makecell[c]{EvoKAN-WF} & \makecell[c]{EvoKAN-SF} & {PINN-SF} \\
			\hline
			{Hidden layers} & {[4, 4, 4]} & \makecell[c]{[4, 4, 4]} & {[15, 15, 15]}\\
            {Activation functions} & \makecell[c]{Gaussian RBFs/SiLU} & \makecell[c]{Gaussian RBFs/SiLU} & {Tanh} \\
            \makecell[l]{Grid points number\\ of activation functions} & {8} & \makecell[c]{5}  & {-}\\
			\makecell[l]{Number of \\ trainable parameters} & {388} & {388} & \makecell[c]{526}\\
            {Optimizer} & \makecell[c]{Adam} & \makecell[c]{Adam} & \makecell[c]{Adam/L-BFGS-B}\\
            {Timestep} & \makecell[c]{1e-03} & \makecell[c]{1e-03} & \makecell[c]{-}\\
			\hline
		\end{tabular}
		\label{table:heat_training}
	\end{center}
\end{table}

Table~\ref{table:1D_AC_training_setting} lists the training configurations
used for EvoKAN-WF, EvoKAN-SF, and PINN-SF.
The two evolutionary KAN models adopt an identical network design
and undergo an initial optimization stage to fit the initial condition
using the Adam optimizer.
After this stage, the solution is advanced in time through parameter updates
with a prescribed fixed time step.
By contrast, PINN-SF relies on a deeper multilayer perceptron with \texttt{tanh} activations
and is trained through standard optimization using Adam followed by L-BFGS-B,
without a time marching mechanism.

\begin{figure}[h!]
    \centering
    \begin{subfigure}[b]{1\linewidth}
        \centering
        \includegraphics[width=\linewidth]{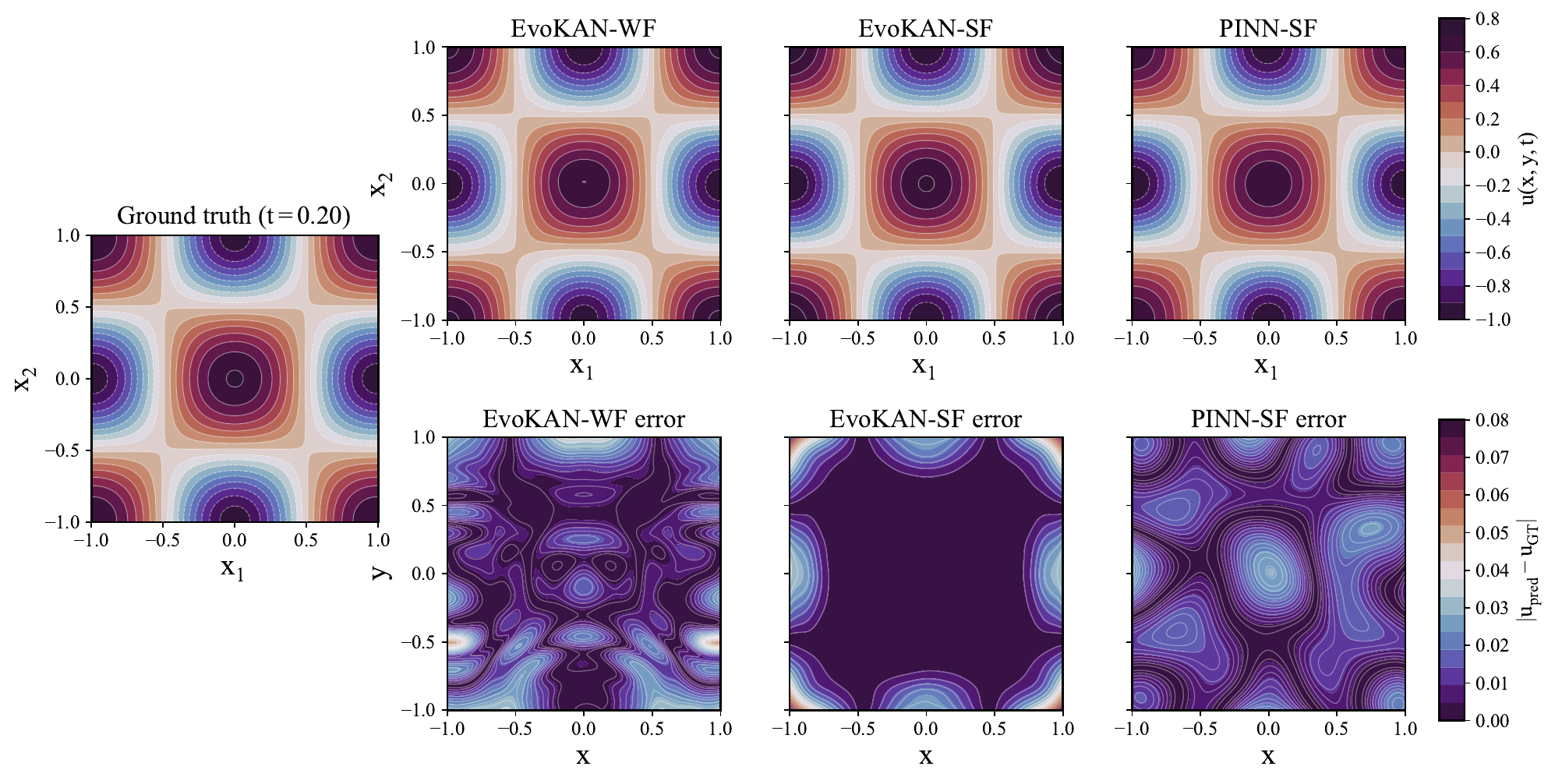}
    \end{subfigure}
\caption{
\textbf{Comparison of two-dimensional heat equation (Eq.~\eqref{eq:Heat_2D_text}) solutions at $t = 0.20$.}
All models are trained using $400$ collocation points.
The top row shows the ground-truth solution together with predictions from EvoKAN-WF, EvoKAN-SF, and PINN-SF.
The bottom row presents the corresponding absolute error fields with respect to the ground truth.
The weak-form solver (EvoKAN-WF) exhibits smoother error distributions and reduced boundary artifacts
compared to the strong-form approaches under the same training budget.
}
\label{fig:2D_heat_solution_02}
\end{figure}

\begin{figure}[h!]
    \centering
    \begin{subfigure}[b]{1\linewidth}
        \centering
        \includegraphics[width=\linewidth]{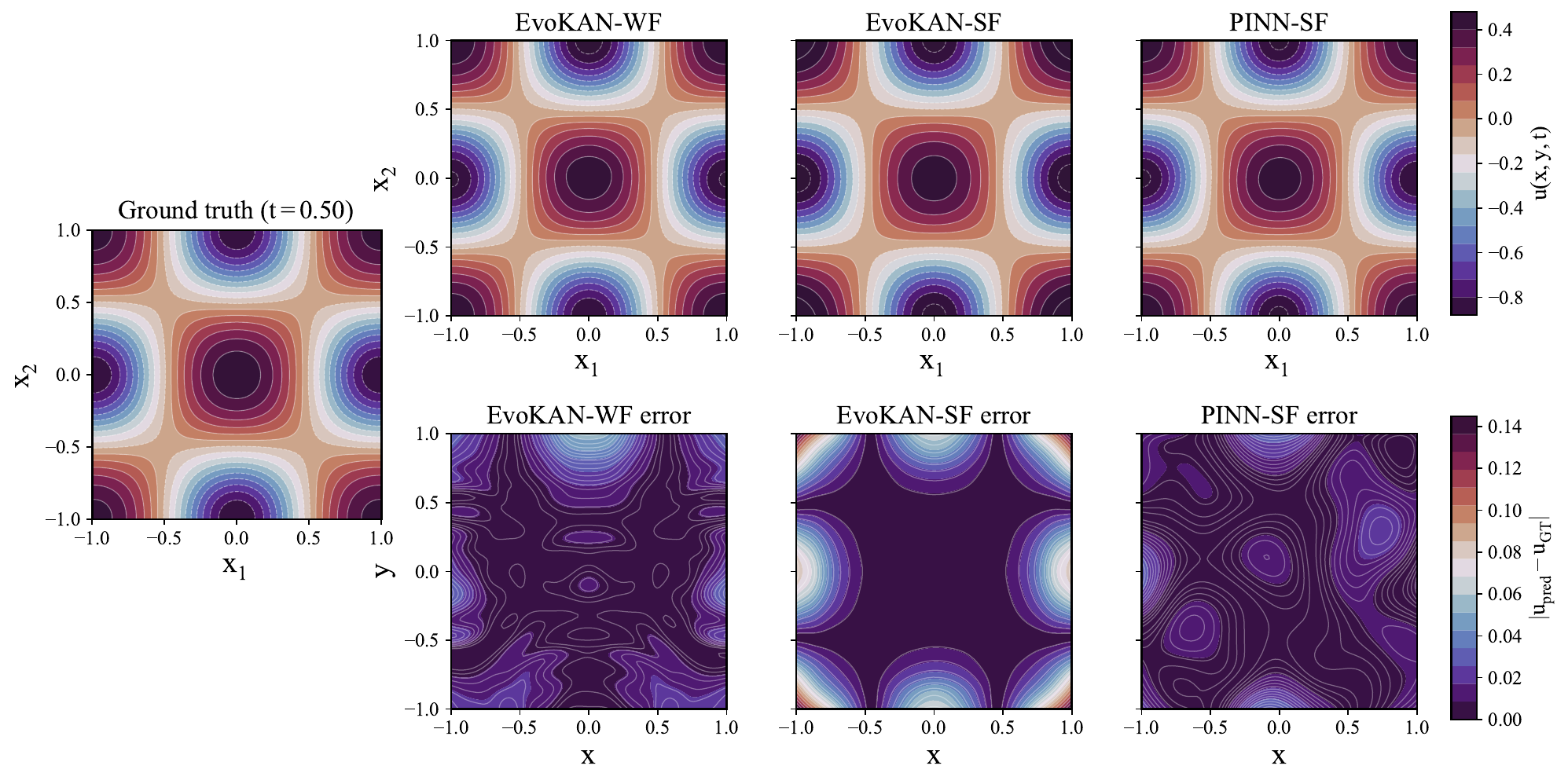}
    \end{subfigure}
\caption{
\textbf{Comparison of two-dimensional heat equation solutions at $t = 0.50$.}
All models are trained using $400$ collocation points.
The top row presents the ground-truth solution along with predictions obtained by EvoKAN-WF, EvoKAN-SF, and PINN-SF.
The bottom row displays the corresponding absolute error fields relative to the ground truth.
Even at this later time, the weak-form solver (EvoKAN-WF) maintains higher accuracy,
exhibiting consistently lower errors and smoother spatial error patterns
than the strong-form approaches under the same training budget.
}
\label{fig:2D_heat_solution_04}
\end{figure}

\begin{figure}[h!]
    \centering
    \begin{subfigure}[b]{0.32\linewidth}
        \centering
        \includegraphics[width=\linewidth]{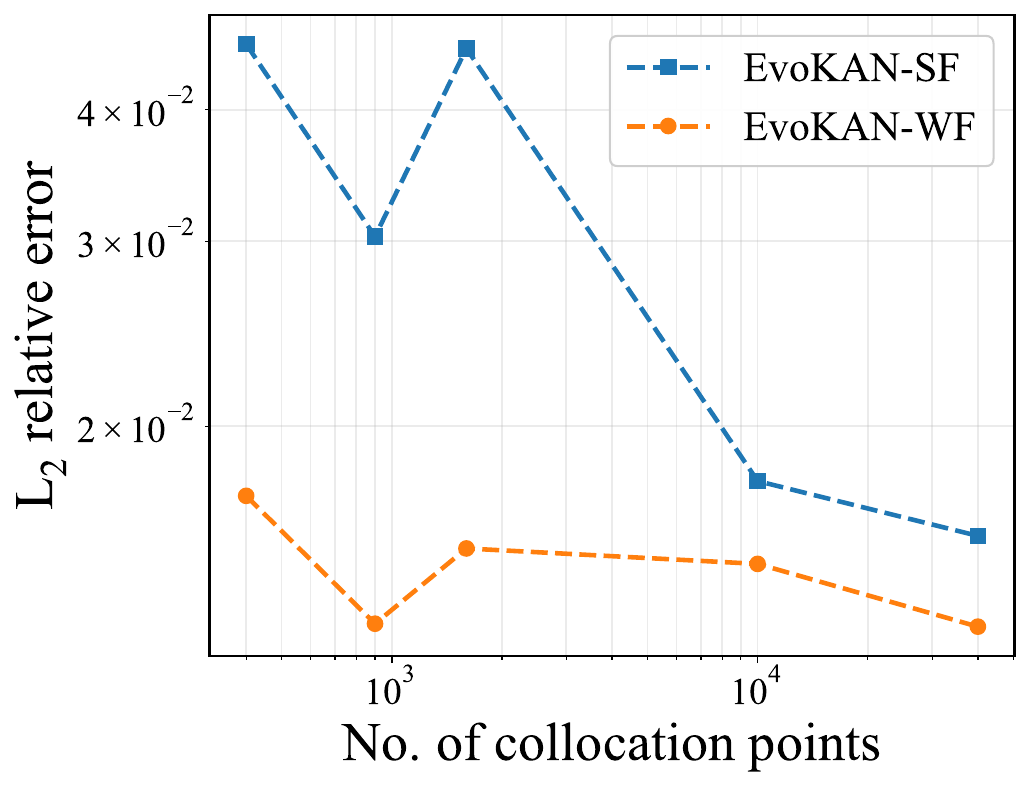}
        \subcaption{$L_2$ relative error}
    \end{subfigure} 
    \begin{subfigure}[b]{0.32\linewidth}
        \centering
        \includegraphics[width=\linewidth]{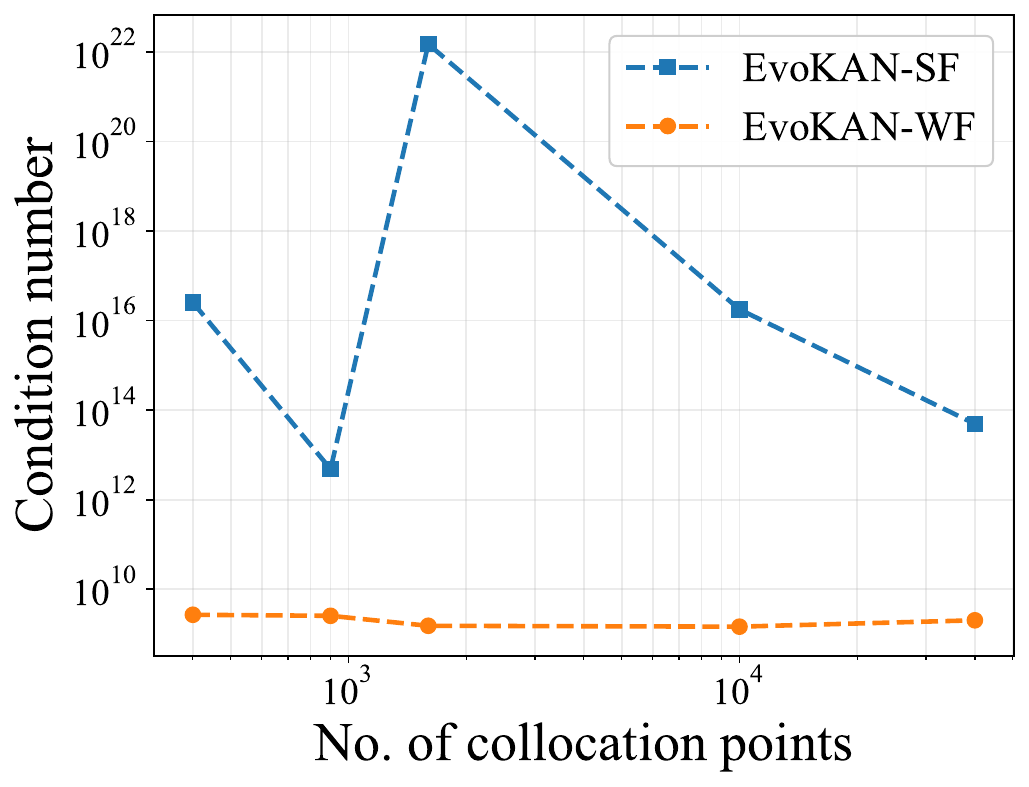}
        \subcaption{Condition number}
    \end{subfigure} 
    \begin{subfigure}[b]{0.32\linewidth}
        \centering
        \includegraphics[width=\linewidth]{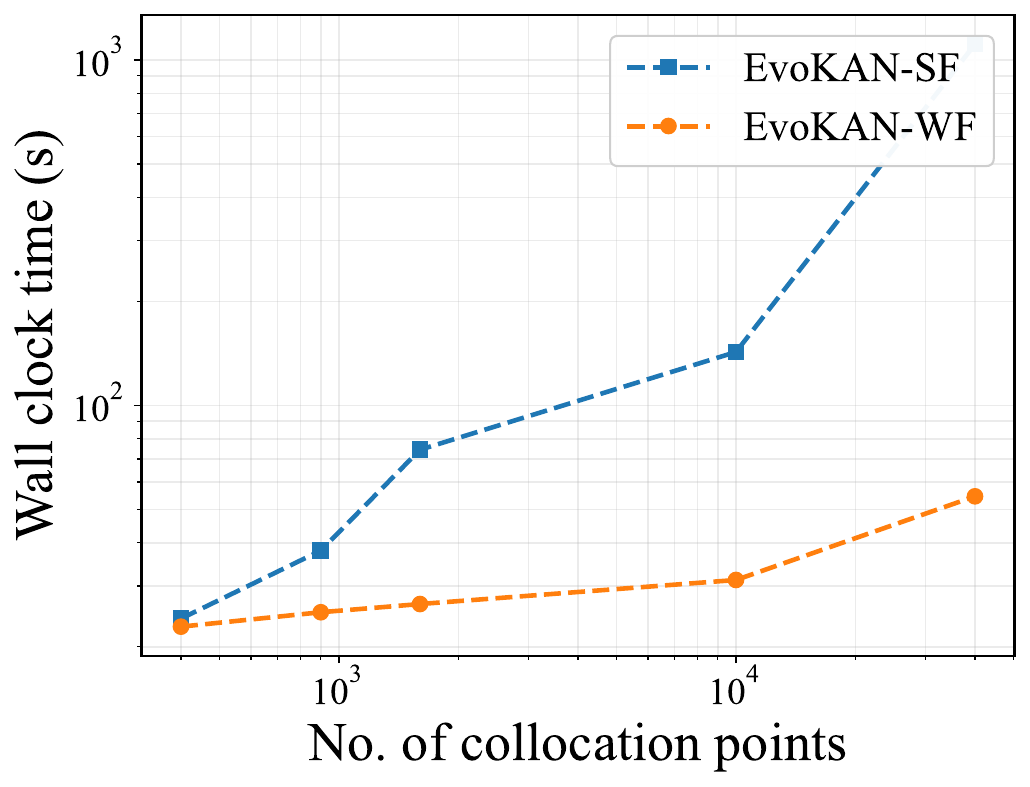}
        \subcaption{Computation time}
    \end{subfigure} 
    \caption{
Strong-form (EvoKAN-SF) versus weak-form (EvoKAN-WF) evolutionary KAN solvers
for the two-dimensional heat equation
with respect to the number of collocation points:
(a) $L_2$ relative error,
(b) condition number,
and (c) wall-clock time.
The weak-form approach achieves lower errors and substantially better conditioning,
while avoiding the rapid growth in computational cost observed in the strong-form solver.
}
\label{fig:2D_heat_summary}
\end{figure}

To examine the solution behavior of different solvers for the two dimensional heat equation,
Figs.~\ref{fig:2D_heat_solution_02} and~\ref{fig:2D_heat_solution_04}
compare the predicted solutions and corresponding absolute error fields
at $t=0.20$ and $t=0.50$ using the same number of collocation points.
At both time instances, EvoKAN WF reproduces the main spatial features of the ground truth
and yields error fields that remain smooth across the domain.
EvoKAN SF shows localized error concentrations near regions with stronger spatial variation,
while PINN SF also captures the overall solution structure and produces error fields
that remain bounded under the same training setting.
As time progresses from $t=0.20$ to $t=0.50$, the differences in error magnitude and spatial structure
remain consistent across the two figures.

The dependence of solution accuracy, conditioning, and computational cost
on the number of collocation points is summarized in Fig.~\ref{fig:2D_heat_summary}
for EvoKAN SF and EvoKAN WF applied to the two dimensional heat equation.
As the number of collocation points increases, EvoKAN SF exhibits variations
in the L2 relative error together with an increase in the condition number.
EvoKAN WF maintains lower error levels and a condition number
that remains bounded across the tested resolutions.
The wall clock time increases with the number of collocation points for EvoKAN SF,
whereas EvoKAN WF shows a weaker dependence on the collocation count.

\begin{figure}[h!]
    \centering
    \begin{subfigure}[b]{0.48\linewidth}
        \centering
        \includegraphics[width=\linewidth]{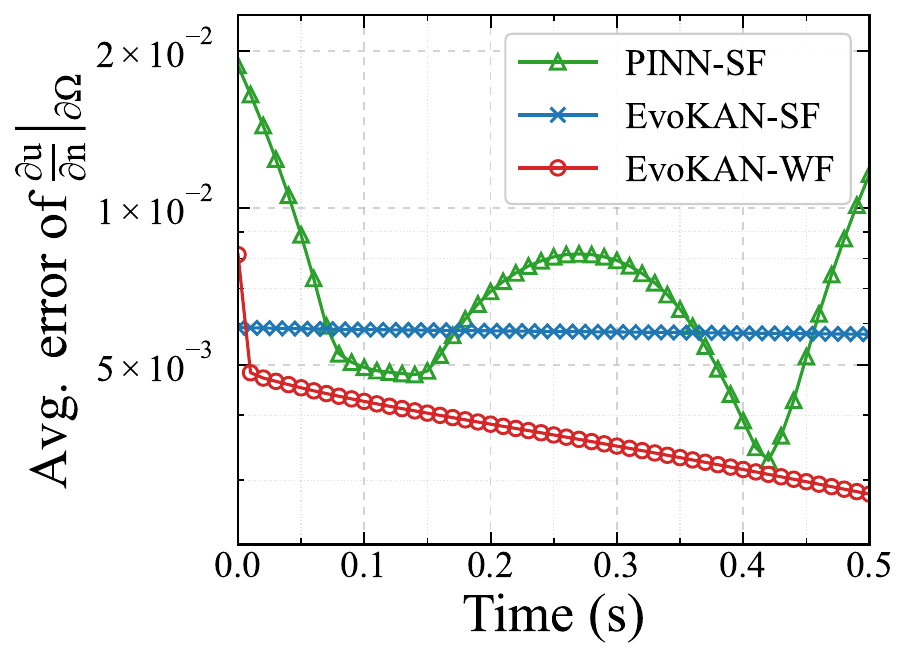}
        \subcaption{Gradient error on the left boundary}
    \end{subfigure}
    \begin{subfigure}[b]{0.48\linewidth}
        \centering
        \includegraphics[width=\linewidth]{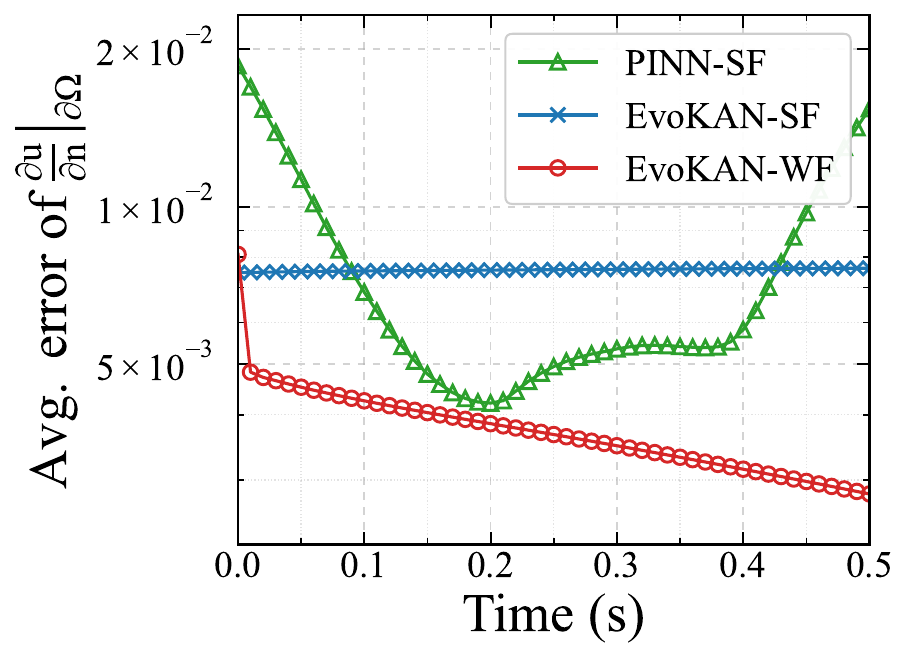}
        \subcaption{Gradient error on the right boundary}
    \end{subfigure} \\
    \begin{subfigure}[b]{0.48\linewidth}
        \centering
        \includegraphics[width=\linewidth]{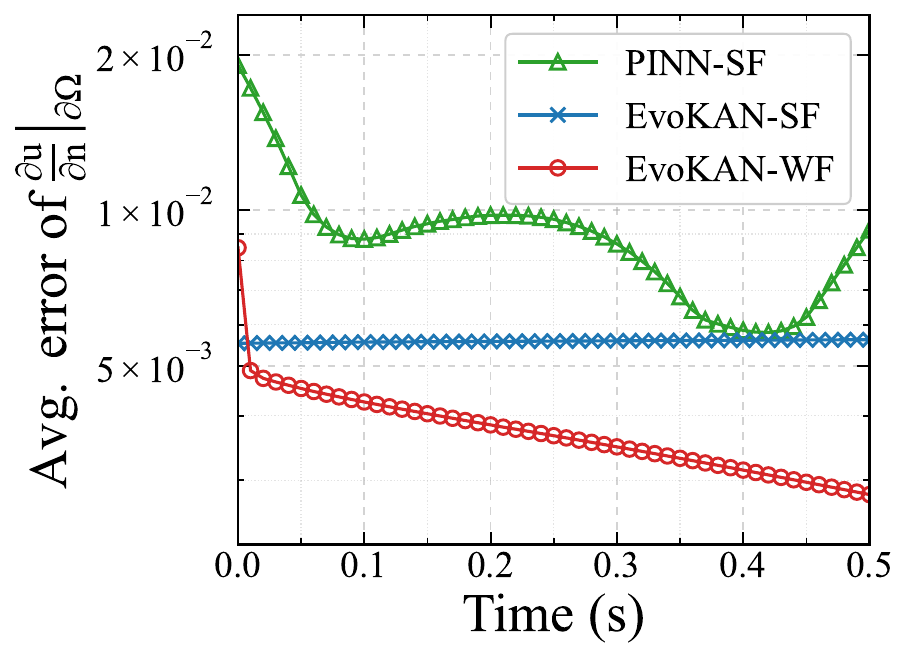}
        \subcaption{Gradient error on the bottom boundary}
    \end{subfigure}
    \begin{subfigure}[b]{0.48\linewidth}
        \centering
        \includegraphics[width=\linewidth]{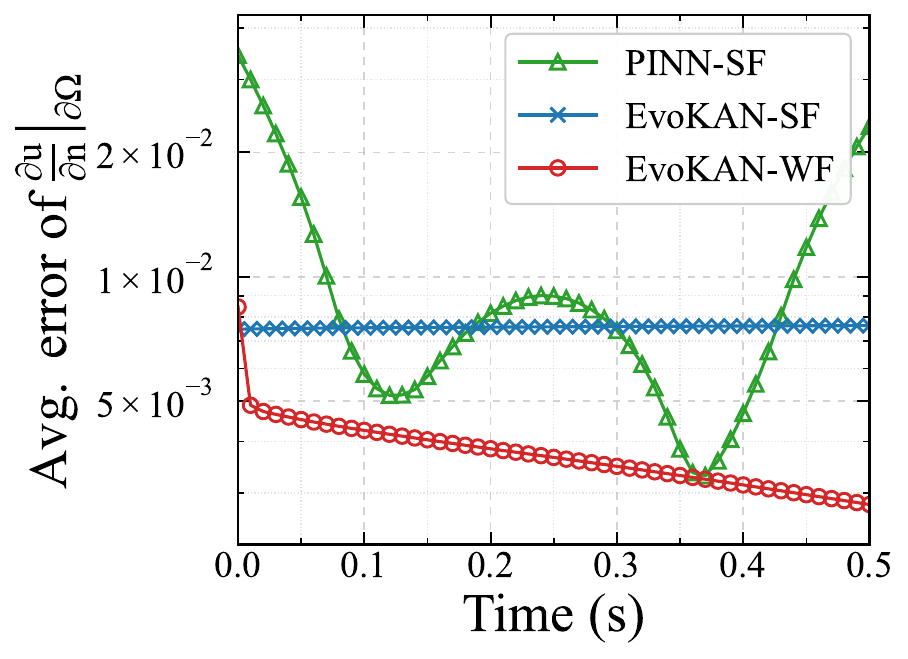}
        \subcaption{Gradient error on the top boundary}
    \end{subfigure}
\caption{
Gradient error on each boundary for the 2D heat equation (Eq.~\eqref{eq:Heat_2D_text}).  
The weak form evolutionary KAN progressively reduces the gradient error as the time evolution proceeds and the error approaches zero along all boundaries.  
The weak formulation provides stable treatment of the Neumann boundary condition and enables monotonic error reduction during network updates.  
The strong-form evolutionary network enforces a constant-derivative boundary condition, making convergence of the boundary error to zero difficult, while the strong-form PINN exhibits oscillatory boundary errors during the time evolution.
}
\label{fig:2d_heat_gradient}
\end{figure}

The evolution of boundary gradient errors over time
for the two dimensional heat equation
is reported in Fig.~\ref{fig:2d_heat_gradient}
for EvoKAN WF, EvoKAN SF, and PINN SF.
Across all four boundaries, EvoKAN WF shows a consistent decrease
in the average gradient error as time progresses.
The error curves remain smooth and follow a monotonic decay pattern
during the time evolution.
EvoKAN SF exhibits slower error reduction
and maintains higher boundary errors over the same time interval.
PINN SF captures the general trend of error decay
but shows larger temporal variations and higher error levels
on all boundaries.
These results indicate that the weak formulation
provides a stable treatment of Neumann boundary conditions
and leads to more controlled boundary error evolution
under the same training setting.

\begin{table}[t]
\footnotesize
\centering
\caption{
$L_2$ relative error and wall-clock time for the two-dimensional heat equation (Eq.~\eqref{eq:Heat_2D_text})
with varying numbers of collocation points.
Speedup is defined as $T_{\text{SF}}/T_{\text{WF}}$.
}
\label{tab:error_time_2d_heat}
\renewcommand{\arraystretch}{1.25}
\begin{tabular}{c cc cc c}
\toprule
& \multicolumn{2}{c}{$L_2$ relative error}
& \multicolumn{3}{c}{Wall-clock time} \\
\cmidrule(lr){2-3} \cmidrule(lr){4-6}
\makecell{Number of \\ collocation points}
& EvoKAN-SF
& EvoKAN-WF
& EvoKAN-SF [s]
& EvoKAN-WF [s]
& Speedup \\
\midrule
400
& 4.6195e-02
& \textbf{1.7171e-02}
& 24.1326
& \textbf{22.8757}
& \textbf{1.06} \\
900
& 3.0303e-02
& \textbf{1.2976e-02}
& 37.9921
& \textbf{25.1838}
& \textbf{1.51} \\
1{,}600
& 4.5767e-02
& \textbf{1.5302e-02}
& 74.4028
& \textbf{26.6005}
& \textbf{2.80} \\
10{,}000
& 1.7737e-02
& \textbf{1.4792e-02}
& 142.7417
& \textbf{31.2263}
& \textbf{4.57} \\
40{,}000
& 1.5728e-02
& \textbf{1.2892e-02}
& 1111.4212
& \textbf{54.5341}
& \textbf{20.39} \\
\bottomrule
\end{tabular}
\end{table}

\subsection{Periodic Boundary Condition: 2D Porous Medium Equation with Drift}

The two-dimensional porous medium equation with drift is written in strong form as
\begin{equation}
\label{eq:PME_drift_strong}
\frac{\partial u}{\partial t}
=
\nabla \cdot \bigl( \nabla (u^2) \bigr)
-
\nabla \cdot \bigl( \mathbf{V}(x,y)\,u \bigr),
\qquad (x,y)\in\Omega,\; t>0,
\end{equation}
with nonlinear diffusion exponent fixed at $m=2$.
The drift velocity field is prescribed by
\begin{equation}
\label{eq:PME_drift_velocity}
\mathbf{V}(x,y)
=
\begin{pmatrix}
\pi \cos(\pi x)\,\sin(\pi y) \\
\pi \sin(\pi x)\,\cos(\pi y)
\end{pmatrix}
=
\nabla\!\bigl( \sin(\pi x)\sin(\pi y) \bigr),
\end{equation}
so that the transport term follows a potential flow.
The computational domain is defined as
\[
\Omega = (-1,1)\times(-1,1).
\]
The problem is equipped with periodic boundary conditions in both spatial directions,
\[
u(-1,y,t)=u(1,y,t),
\qquad
u(x,-1,t)=u(x,1,t),
\]
together with periodicity of the associated fluxes.

To derive the weak formulation, a family of test functions $\{v_k\}_{k=1}^K$ is introduced.
Each test function is defined using trigonometric Fourier modes of the form
$\cos(m\pi x)$ or $\sin(m\pi x)$ with mode index $m=1,2,\ldots$ in each spatial direction.
The mode indices are chosen to satisfy periodicity on $\partial\Omega$,
and the constant mode is excluded to avoid redundancy in the test space.
Multiplying \eqref{eq:PME_drift_strong} by $v_k$ and integrating over $\Omega$ yields
\begin{equation}
\int_{\Omega}
\left(
\frac{\partial u}{\partial t}
-
\nabla\cdot\bigl(\nabla(u^2)\bigr)
+
\nabla\cdot\bigl(\mathbf{V}u\bigr)
\right)
v_k
\,dx\,dy
=0,
\qquad k=1,\ldots,K.
\end{equation}
The divergence terms are treated by integration by parts.
For the nonlinear diffusion term,
\[
-\int_{\Omega}
\nabla\cdot\bigl(\nabla(u^2)\bigr)\,v_k\,dx\,dy
=
\int_{\Omega}
\nabla(u^2)\cdot\nabla v_k\,dx\,dy
-
\int_{\partial\Omega}
\nabla(u^2)\cdot\mathbf{n}\,v_k\,ds,
\]
and for the drift term,
\[
\int_{\Omega}
\nabla\cdot\bigl(\mathbf{V}u\bigr)\,v_k\,dx\,dy
=
-\int_{\Omega}
(\mathbf{V}u)\cdot\nabla v_k\,dx\,dy
+
\int_{\partial\Omega}
(\mathbf{V}u)\cdot\mathbf{n}\,v_k\,ds,
\]
where $\mathbf{n}$ denotes the outward unit normal on $\partial\Omega$.
The boundary integrals cancel under periodic boundary conditions
because opposite sides of the domain contribute equal and opposite flux terms.
Substituting these relations into the integral balance yields the weak form
\begin{equation}
\label{eq:PME_drift_weak}
\int_{\Omega}
\frac{\partial u}{\partial t}\,v_k\,dx\,dy
=
-\int_{\Omega}
\nabla(u^2)\cdot\nabla v_k\,dx\,dy
+
\int_{\Omega}
(\mathbf{V}u)\cdot\nabla v_k\,dx\,dy,
\qquad k=1,\ldots,K.
\end{equation}
Equation~\eqref{eq:PME_drift_weak} defines the weak formulation of the porous medium equation
with drift under periodic boundary conditions
and serves as the spatial weak form for the numerical experiments.

\sisetup{group-separator={,}, group-minimum-digits=4}
\begin{table} [hbt!]
\footnotesize
	\renewcommand{\arraystretch}{1.0}
	\begin{center} 
		\caption{Training configuration for the 2D porous medium equation with drift (Eq.~\eqref{eq:PME_drift_strong}).}
		\begin{tabular}{l c c c}
			\hline
			{\, \, \, } & \makecell[c]{EvoKAN-WF} & \makecell[c]{EvoKAN-SF} & {PINN-SF} \\
			\hline
			{Hidden layers} & {[7, 7, 7]} & \makecell[c]{[7, 7, 7]} & {[15, 15, 15]}\\
            {Activation functions} & \makecell[c]{Gaussian RBFs/SiLU} & \makecell[c]{Gaussian RBFs/SiLU} & {Tanh} \\
            \makecell[l]{Grid points number\\ of activation functions} & {5} & \makecell[c]{5}  & {-}\\
			\makecell[l]{Number of \\ trainable parameters} & {770} & {770} & \makecell[c]{526}\\
            {Optimizer} & \makecell[c]{Adam} & \makecell[c]{Adam} & \makecell[c]{Adam/L-BFGS-B}\\
            {Timestep} & \makecell[c]{1e-03} & \makecell[c]{1e-03} & \makecell[c]{-}\\
			\hline
		\end{tabular}
		\label{table:PME_training}
	\end{center}
\end{table}

\begin{figure}[h!]
    \centering
        \begin{subfigure}[b]{1\linewidth}
        \centering
        \includegraphics[width=\linewidth]{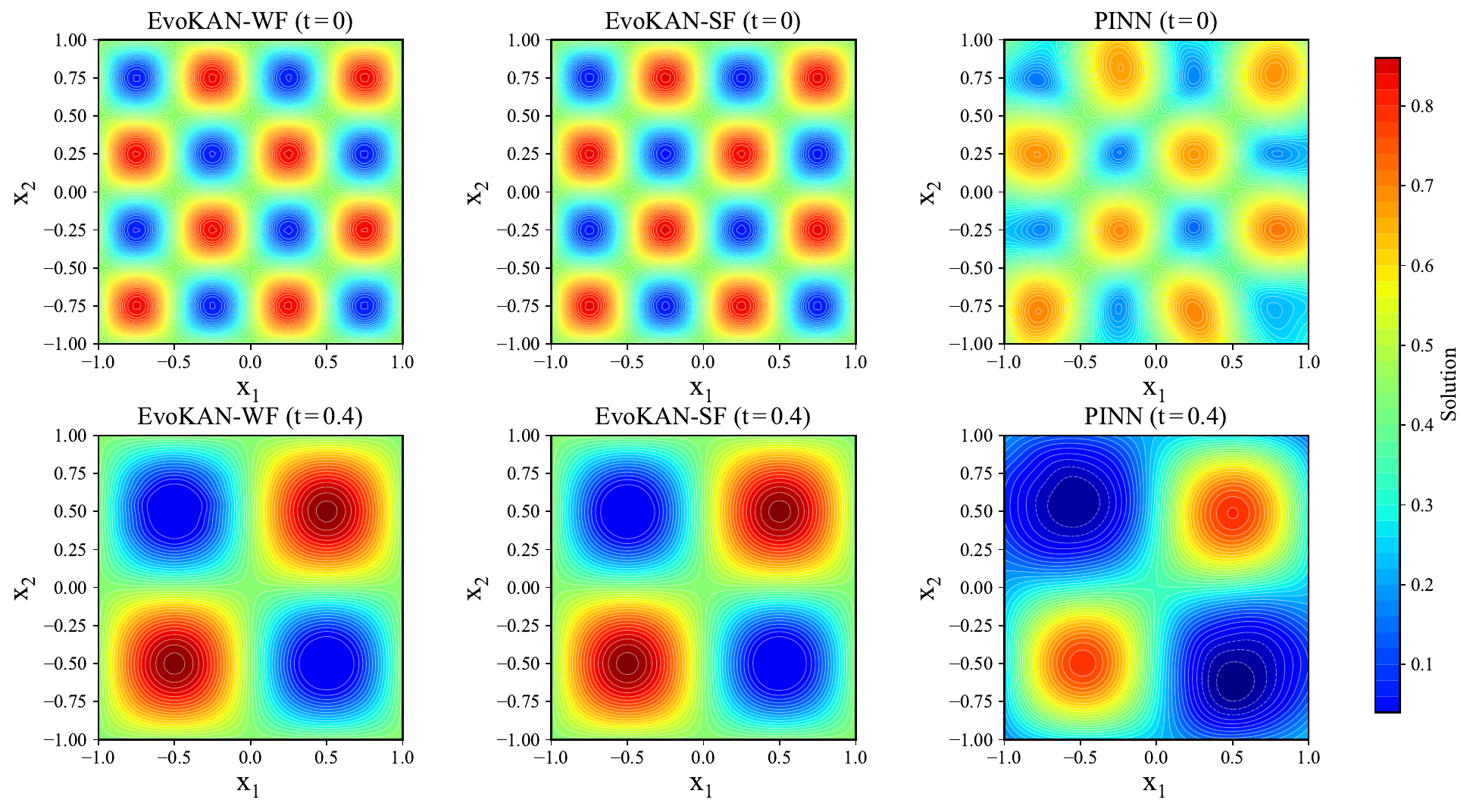}
    \end{subfigure}
    \caption{
\textbf{Solution field comparison for the two-dimensional PME with drift under periodic boundary conditions~(Eq.~\eqref{eq:PME_drift_strong}).}
The solution fields at $t=0$ (top row) and $t=0.4$ (bottom row) are shown for EvoKAN-WF, EvoKAN-SF, and PINN, respectively.
All models are trained using $N_c = 3600$ collocation points, with a uniform time step $\Delta t = 10^{-2}$ up to the final time $T = 0.4$.
The color map represents the spatial distribution of the solution $u(x_1,x_2,t)$, illustrating the nonlinear diffusion and drift-induced transport dynamics.
}
\label{fig:PME_solution}
\end{figure}

The solution fields for the two dimensional porous medium equation with drift
under periodic boundary conditions are shown in Fig.~\ref{fig:PME_solution}.
At the initial time $t=0$, all methods reproduce the prescribed initial distribution.
At the final time $t=0.4$, EvoKAN-WF, EvoKAN-SF, and PINN capture the large scale transport patterns
induced by the combined nonlinear diffusion and drift.
The spatial structures remain periodic across the domain,
reflecting consistent enforcement of the periodic boundary conditions.
Under the same number of collocation points and time step,
the predicted solution fields exhibit comparable global behavior
while differing in the smoothness of local variations.

\begin{figure}[h!]
    \centering
    \begin{subfigure}[b]{0.45\linewidth}
        \centering
        \includegraphics[width=\linewidth]{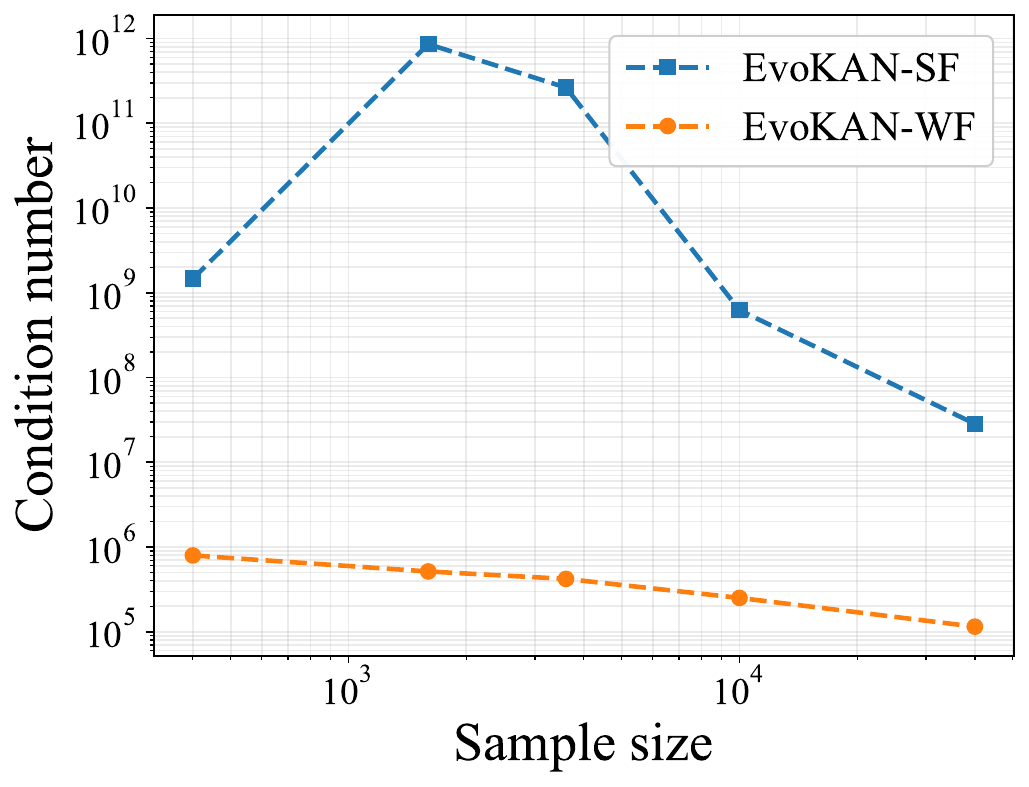}
        \subcaption{Condition number}
    \end{subfigure} 
        \begin{subfigure}[b]{0.45\linewidth}
        \centering
        \includegraphics[width=\linewidth]{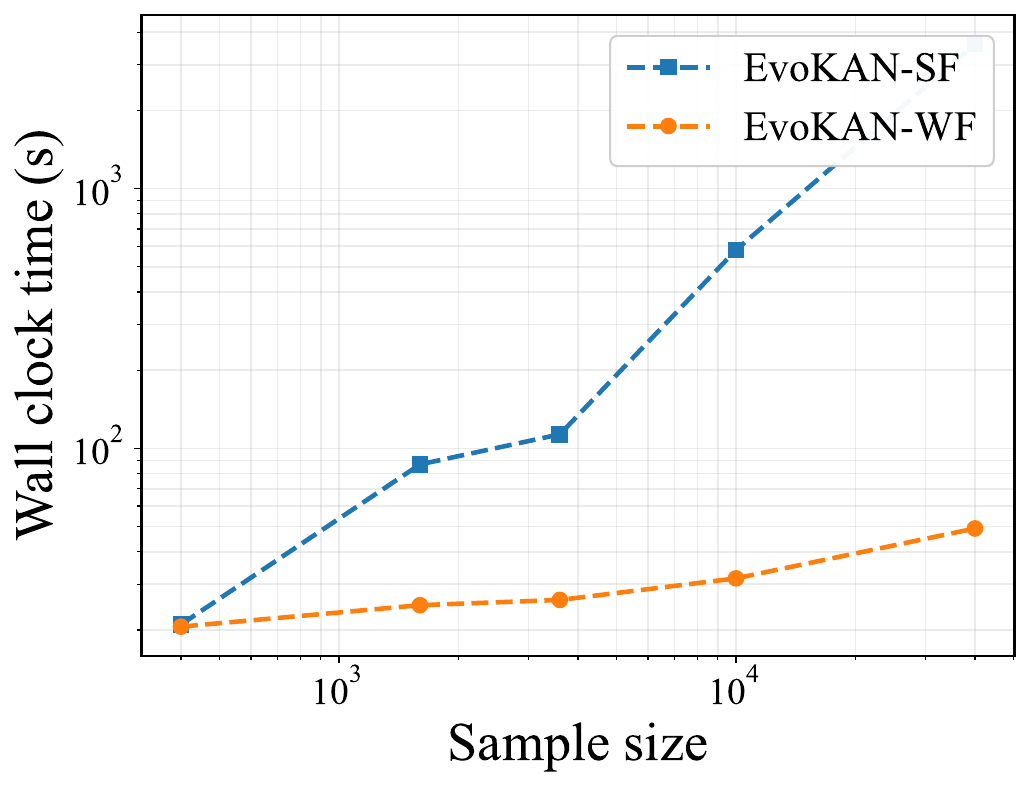}
        \subcaption{Computation time}
    \end{subfigure}
        \caption{
Strong-form (EvoKAN-SF) versus weak-form (EvoKAN-WF) evolutionary KAN solvers
for the two-dimensional porous medium equation (Eq.~\eqref{eq:PME_drift_strong})
with respect to the number of collocation points:
(a) condition number and
(b) wall-clock time.
The weak-form approach achieves substantially better conditioning
while avoiding the rapid growth in computational cost observed in the strong-form solver.
}
\label{fig:PME_summary}
\end{figure}

The dependence of conditioning and computational cost on the number of collocation points
for the two dimensional porous medium equation is shown in Fig.~\ref{fig:PME_summary}.
As the sample size increases, EvoKAN-SF exhibits a pronounced increase in the condition number
together with a rapid growth in wall clock time.
In contrast, EvoKAN-WF maintains controlled conditioning
and a gradual increase in computational cost.
These results indicate that the weak form formulation
leads to more stable parameter update systems
and more favorable computational scaling
than the strong form approach for this problem.

\begin{figure}[h!]
    \centering
    \begin{subfigure}[b]{0.32\linewidth}
        \centering
        \includegraphics[width=\linewidth]{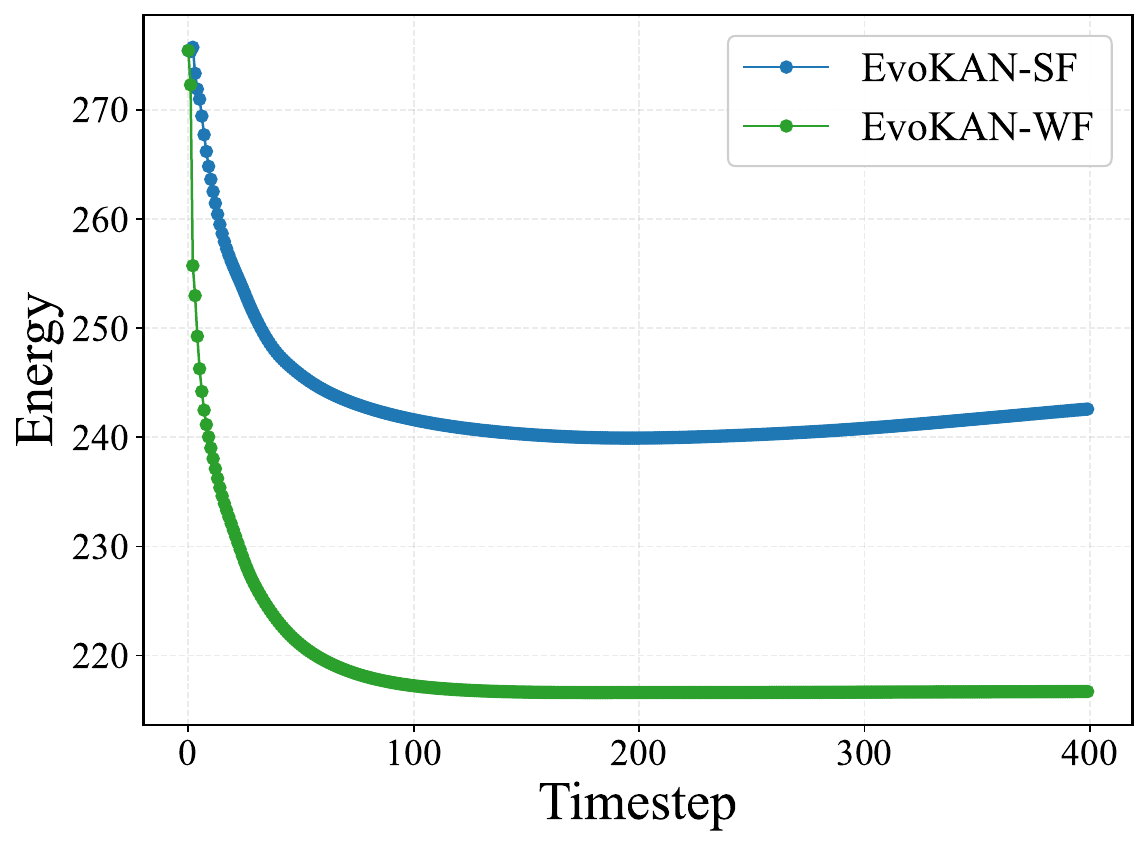}
        \subcaption{Collocation points: 400}
    \end{subfigure} 
    \begin{subfigure}[b]{0.32\linewidth}
        \centering
        \includegraphics[width=\linewidth]{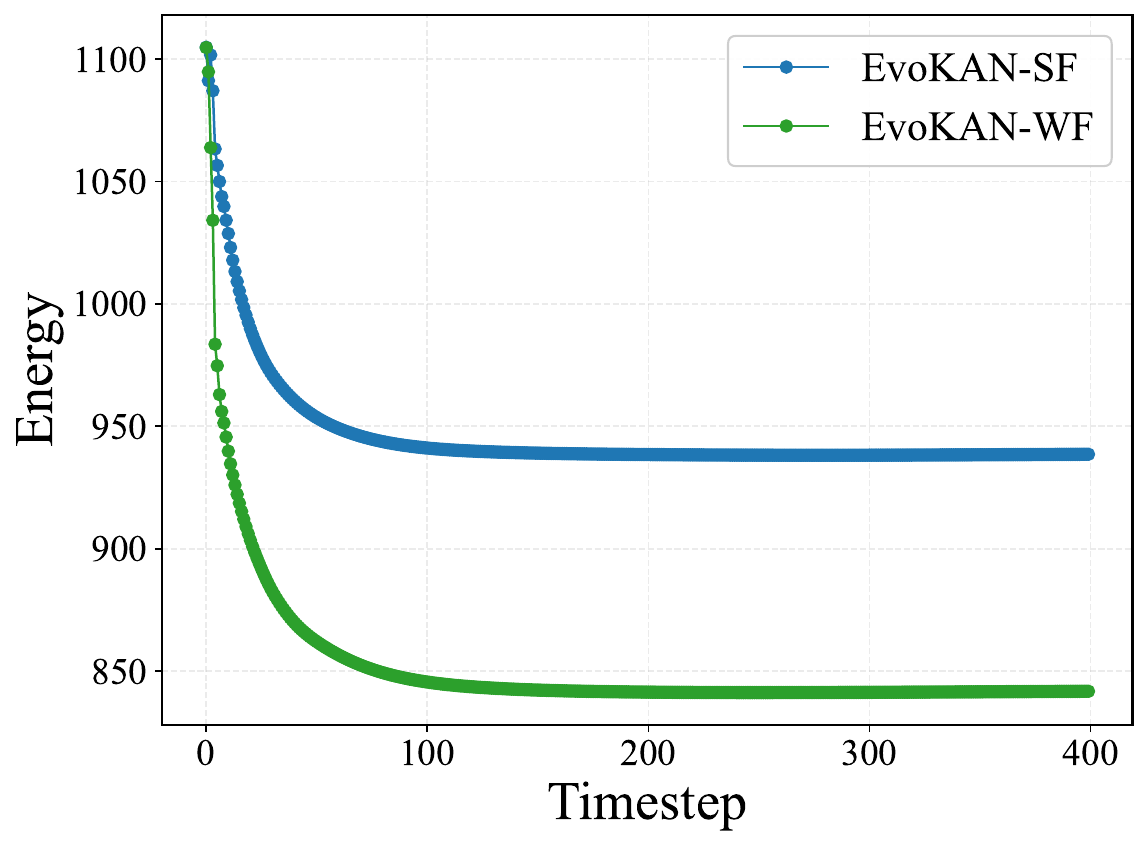}
        \subcaption{Collocation points: 1600}
    \end{subfigure}
    \begin{subfigure}[b]{0.32\linewidth}
        \centering
        \includegraphics[width=\linewidth]{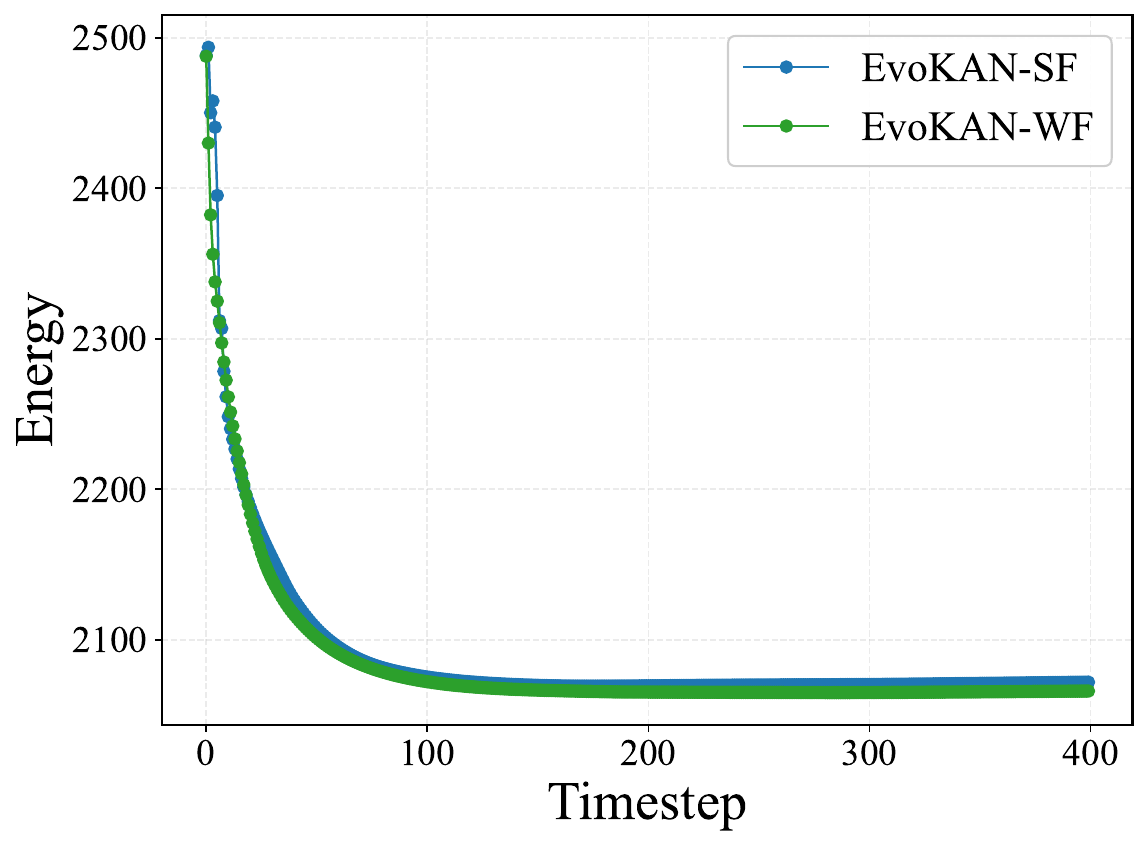}
        \subcaption{Collocation points: 3600}
    \end{subfigure}
\caption{
Energy evolution of strong-form (EvoKAN-SF) and weak-form (EvoKAN-WF)
evolutionary KAN solvers for the two-dimensional porous medium equation
under different numbers of collocation points.
For EvoKAN-SF, monotone energy decay is observed only when a sufficiently large number
of collocation points is used.
At lower resolutions, EvoKAN-SF exhibits energy increase
or incomplete decay over time.
In contrast, EvoKAN-WF shows consistent energy decay
across all tested resolutions.
}

\label{fig:PME_energy}
\end{figure}

The energy evolution for the two dimensional porous medium equation
is shown in Fig.~\ref{fig:PME_energy} for different numbers of collocation points.
In the case of $400$ collocation points, EvoKAN-SF exhibits a temporary increase in energy,
indicating that the solution does not follow the expected dissipative behavior.
For $1600$ collocation points, the energy decay of EvoKAN-SF remains incomplete
over the simulated time interval.
Only when $3600$ collocation points are employed
does EvoKAN-SF exhibit monotone energy decay.
In contrast, EvoKAN-WF shows consistent energy decay
for all tested resolutions.
This behavior indicates that the weak form formulation
preserves the dissipative structure of the porous medium equation
with reduced sensitivity to the number of collocation points.

\begin{table}[t]
\footnotesize
\centering
\caption{
Solution validity and wall clock time for the 2D PME (Eq.~\eqref{eq:PME_drift_strong})
with varying the number of collocation points.
The solution validity indicator reports whether the numerical solution
exhibits a dissipative energy evolution consistent with the governing equation.
Entries marked with $\times$ correspond to cases where the computed energy
shows non-decreasing behavior or incomplete decay over the simulated time interval,
indicating that the solution does not provide an admissible approximation
under the given discretization.
Speedup is defined as $T_{\text{SF}}/T_{\text{WF}}$.
}
\label{tab:energy_time_pme}
\renewcommand{\arraystretch}{1.25}
\begin{tabular}{c cc cc c}
\toprule
& \multicolumn{2}{c}{Approximation validity}
& \multicolumn{3}{c}{Wall clock time} \\
\cmidrule(lr){2-3} \cmidrule(lr){4-6}
\makecell{Number of \\ collocation points}
& EvoKAN-SF
& EvoKAN-WF
& EvoKAN-SF [s]
& EvoKAN-WF [s]
& Speedup \\
\midrule
400
& $\times$
& $\bigcirc$
& 21.0145
& \textbf{20.6104}
& \textbf{1.02} \\
1{,}600
& $\times$
& $\bigcirc$
& 86.8148
& \textbf{24.9313}
& \textbf{3.48} \\
3{,}600
& $\bigcirc$
& $\bigcirc$
& 113.3219
& \textbf{26.1163}
& \textbf{4.34} \\
10{,}000
& $\bigcirc$
& $\bigcirc$
& 580.5136
& \textbf{31.6187}
& \textbf{18.37} \\
40{,}000
& $\bigcirc$
& $\bigcirc$
& 3599.2407
& \textbf{49.2156}
& \textbf{73.12} \\
\bottomrule
\end{tabular}
\label{table:PME_accuracy_and_time}
\end{table}

The solution validity and computational cost
are summarized in Table~\ref{table:PME_accuracy_and_time}.
For smaller numbers of collocation points,
EvoKAN-SF fails to produce a valid solution,
as indicated by non-decreasing or incomplete energy decay,
whereas EvoKAN-WF yields admissible solutions in all tested cases.
As the number of collocation points increases,
both methods produce valid solutions,
but the wall clock time of EvoKAN-SF increases rapidly.
In contrast, EvoKAN-WF maintains a lower computational cost,
leading to increasing speedup as the sample size grows.

\section{Conclusion}
\label{sec:conclusion}

This work presents a weak-form evolutionary Kolmogorov--Arnold network framework
for solving \ac{PDEs}.
The main contributions of this work are threefold:
(i) the proposed formulation constructs parameter update systems whose size is
fixed by the number of test functions and remains independent of the number of
training samples, enabling scalable computational efficiency;
(ii) by replacing pointwise residual evaluation with weak residual projections
and reducing the order of differentiation, the method produces well-conditioned
linear systems and maintains accuracy for solutions approaching discontinuities;
and
(iii) Dirichlet and periodic boundary conditions are enforced through
boundary-constrained trial spaces, while Neumann conditions are incorporated
directly into the weak formulation via integration by parts.

The numerical results demonstrate consistent differences between weak-form and
strong-form evolutionary solvers across multiple benchmark problems.
For the one-dimensional Allen--Cahn equation
(Figs.~\ref{fig:1D_AC_solution_1}, \ref{fig:1D_AC_solution_2} and Table~\ref{table:1D_AC_error_time}),
the weak-form solver maintained accurate interface evolution and bounded condition
numbers even with very few collocation points, while the strong-form solver exhibited
accuracy degradation and ill-conditioning at low resolutions.
For the two-dimensional heat equation with Neumann boundary conditions
(Figs.~\ref{fig:2D_heat_solution_02}--\ref{fig:2D_heat_summary}),
the weak-form formulation produced lower $L_2$ errors and smoother error fields
under the same training budget, and enabled stable reduction of boundary gradient
errors during time evolution (Fig.~\ref{fig:2d_heat_gradient}).

In the two-dimensional Burgers' equation
(Figs.~\ref{fig:2D_Burgers_solution_01}--\ref{fig:2D_Burgers_summary}),
the weak-form evolutionary solver preserved the global flow structure under limited
collocation data and achieved lower errors than strong-form evolutionary solvers.
As the number of collocation points increased, the computational cost of the
strong-form solver grew rapidly, whereas the weak-form solver showed weak dependence
on the sample size.
For the two-dimensional porous medium equation with drift
(Figs.~\ref{fig:PME_energy} and Table~\ref{table:PME_accuracy_and_time}),
the weak-form solver consistently produced solutions with dissipative energy
evolution across all tested resolutions, while the strong-form solver required
sufficiently large collocation sets to recover admissible energy decay.

Overall, the weak-form evolutionary framework enables accurate resolution of
complex physical phenomena with controlled conditioning and stable time
evolution, even under limited computational resources.
Future work will investigate uncertainty quantification and applications to
practical engineering systems, extending the present framework toward
realistic, data-constrained physical problems.








\section*{Acknowledgment}
We would like to thank the support of National Science Foundation (DMS-2533878, DMS-2053746, DMS-2134209, ECCS-2328241, CBET-2347401 and OAC-2311848), and U.S.~Department of Energy (DOE) Office of Science Advanced Scientific Computing Research program DE-SC0023161, the SciDAC LEADS Institute, and DOE–Fusion Energy Science, under grant number: DE-SC0024583.

\bibliographystyle{unsrt}
\bibliography{reference}

@article{zhang2024energy,
  title={Energy-dissipative evolutionary deep operator neural networks},
  author={Zhang, Jiahao and Zhang, Shiheng and Shen, Jie and Lin, Guang},
  journal={Journal of Computational Physics},
  volume={498},
  pages={112638},
  year={2024},
  publisher={Elsevier}
}

@article{liu2024kan2,
  title={Kan 2.0: Kolmogorov-arnold networks meet science},
  author={Liu, Ziming and Ma, Pingchuan and Wang, Yixuan and Matusik, Wojciech and Tegmark, Max},
  journal={arXiv preprint arXiv:2408.10205},
  year={2024}
}

@book{kolmogorov1961representation,
  title={On the representation of continuous functions of several variables by superpositions of continuous functions of a smaller number of variables},
  author={Kolmogorov, Andre{{\u\i}} Nikolaevich},
  year={1961},
  publisher={American Mathematical Society}
}

@inproceedings{kolmogorov1957representation,
  title={On the representation of continuous functions of many variables by superposition of continuous functions of one variable and addition},
  author={Kolmogorov, Andrei Nikolaevich},
  booktitle={Doklady Akademii Nauk},
  volume={114},
  number={5},
  pages={953--956},
  year={1957},
  organization={Russian Academy of Sciences}
}

@article{braun2009constructive,
  title={On a constructive proof of Kolmogorov’s superposition theorem},
  author={Braun, J{\"u}rgen and Griebel, Michael},
  journal={Constructive approximation},
  volume={30},
  pages={653--675},
  year={2009},
  publisher={Springer}
}

@article{abueidda2024deepokan,
  title={Deepokan: Deep operator network based on kolmogorov arnold networks for mechanics problems},
  author={Abueidda, Diab W and Pantidis, Panos and Mobasher, Mostafa E},
  journal={arXiv preprint arXiv:2405.19143},
  year={2024}
}

@article{cai2021physics,
  title={Physics-informed neural networks (PINNs) for fluid mechanics: A review},
  author={Cai, Shengze and Mao, Zhiping and Wang, Zhicheng and Yin, Minglang and Karniadakis, George Em},
  journal={Acta Mechanica Sinica},
  volume={37},
  number={12},
  pages={1727--1738},
  year={2021},
  publisher={Springer}
}

@article{goswami2022physics,
  title={A physics-informed variational DeepONet for predicting crack path in quasi-brittle materials},
  author={Goswami, Somdatta and Yin, Minglang and Yu, Yue and Karniadakis, George Em},
  journal={Computer Methods in Applied Mechanics and Engineering},
  volume={391},
  pages={114587},
  year={2022},
  publisher={Elsevier}
}

@article{berrone2023enforcing,
  title={Enforcing Dirichlet boundary conditions in physics-informed neural networks and variational physics-informed neural networks},
  author={Berrone, Stefano and Canuto, Claudio and Pintore, Moreno and Sukumar, Natarajan},
  journal={Heliyon},
  volume={9},
  number={8},
  year={2023},
  publisher={Elsevier}
}

@article{liu2024kan,
  title={Kan: Kolmogorov-arnold networks},
  author={Liu, Ziming and Wang, Yixuan and Vaidya, Sachin and Ruehle, Fabian and Halverson, James and Solja{\v{c}}i{\'c}, Marin and Hou, Thomas Y and Tegmark, Max},
  journal={arXiv preprint arXiv:2404.19756},
  year={2024}
}

@misc{orr1996introduction,
  title={Introduction to radial basis function networks},
  author={Orr, Mark JL and others},
  year={1996},
  publisher={Technical Report, center for cognitive science, University of Edinburgh~…}
}

@article{buhmann2000radial,
  title={Radial basis functions},
  author={Buhmann, Martin Dietrich},
  journal={Acta numerica},
  volume={9},
  pages={1--38},
  year={2000},
  publisher={Cambridge university press}
}

@article{li2405kolmogorov,
  title={Kolmogorov-arnold networks are radial basis function networks. arXiv 2024},
  author={Li, Z},
  journal={arXiv preprint arXiv:2405.06721},
  year={2024}
}

@article{lin2025energy,
  title={Energy-dissipative evolutionary kolmogorov-arnold networks for complex pde systems},
  author={Lin, Guang and Mou, Changhong and Zhang, Jiahao},
  journal={arXiv preprint arXiv:2503.01618},
  year={2025}
}

@article{he2023optimal,
  title={On the optimal expressive power of relu dnns and its application in approximation with kolmogorov superposition theorem},
  author={He, Juncai},
  journal={arXiv preprint arXiv:2308.05509},
  year={2023}
}

@incollection{leni2013kolmogorov,
  title={The kolmogorov spline network for image processing},
  author={Leni, Pierre-Emmanuel and Fougerolle, Yohan D and Truchetet, Fr{\'e}d{\'e}ric},
  booktitle={Image Processing: Concepts, Methodologies, Tools, and Applications},
  pages={54--78},
  year={2013},
  publisher={IGI Global}
}

@article{lai2021kolmogorov,
  title={The kolmogorov superposition theorem can break the curse of dimensionality when approximating high dimensional functions},
  author={Lai, Ming-Jun and Shen, Zhaiming},
  journal={arXiv preprint arXiv:2112.09963},
  year={2021}
}

@article{lin1993realization,
  title={On the realization of a Kolmogorov network},
  author={Lin, Ji-Nan and Unbehauen, Rolf},
  journal={Neural Computation},
  volume={5},
  number={1},
  pages={18--20},
  year={1993},
  publisher={MIT Press}
}

@inproceedings{koppen2002training,
  title={On the training of a Kolmogorov Network},
  author={K{\"o}ppen, Mario},
  booktitle={Artificial Neural Networks ICANN 2002: International Conference Madrid, Spain, August 28--30, 2002 Proceedings 12},
  pages={474--479},
  year={2002},
  organization={Springer}
}

@article{fakhoury2022exsplinet,
  title={ExSpliNet: An interpretable and expressive spline-based neural network},
  author={Fakhoury, Daniele and Fakhoury, Emanuele and Speleers, Hendrik},
  journal={Neural Networks},
  volume={152},
  pages={332--346},
  year={2022},
  publisher={Elsevier}
}

@article{yu2018deep,
  title={The deep Ritz method: a deep learning-based numerical algorithm for solving variational problems},
  author={Yu, Bing and others},
  journal={Communications in Mathematics and Statistics},
  volume={6},
  number={1},
  pages={1--12},
  year={2018},
  publisher={Springer}
}

@article{du2021evolutional,
  title={Evolutional deep neural network},
  author={Du, Yifan and Zaki, Tamer A},
  journal={Physical Review E},
  volume={104},
  number={4},
  pages={045303},
  year={2021},
  publisher={APS}
}

@article{apicella2021survey,
  title={A survey on modern trainable activation functions},
  author={Apicella, Andrea and Donnarumma, Francesco and Isgr{\`o}, Francesco and Prevete, Roberto},
  journal={Neural Networks},
  volume={138},
  pages={14--32},
  year={2021},
  publisher={Elsevier}
}

@article{trentin2001networks,
  title={Networks with trainable amplitude of activation functions},
  author={Trentin, Edmondo},
  journal={Neural Networks},
  volume={14},
  number={4-5},
  pages={471--493},
  year={2001},
  publisher={Elsevier}
}

@article{sze2017efficient,
  title={Efficient processing of deep neural networks: A tutorial and survey},
  author={Sze, Vivienne and Chen, Yu-Hsin and Yang, Tien-Ju and Emer, Joel S},
  journal={Proceedings of the IEEE},
  volume={105},
  number={12},
  pages={2295--2329},
  year={2017},
  publisher={Ieee}
}

@article{sirignano2018dgm,
  title     = {DGM: A deep learning algorithm for solving partial differential equations},
  author    = {Sirignano, Justin and Spiliopoulos, Konstantinos},
  journal   = {Journal of Computational Physics},
  volume    = {375},
  pages     = {1339--1364},
  year      = {2018}
}

@article{raissi2019pinn,
  title     = {Physics-informed neural networks: A deep learning framework for solving forward and inverse problems involving nonlinear partial differential equations},
  author    = {Raissi, Maziar and Perdikaris, Paris and Karniadakis, George},
  journal   = {Journal of Computational Physics},
  volume    = {378},
  pages     = {686--707},
  year      = {2019}
}

@article{lu2021deeponet,
  title   = {Learning nonlinear operators via DeepONet based on the universal approximation theorem of operators},
  author  = {Lu, Lu and Jin, Pengzhan and Karniadakis, George Em},
  journal = {Nature Machine Intelligence},
  volume  = {3},
  pages   = {218--229},
  year    = {2021}
}

@inproceedings{li2021fourier,
  title     = {Fourier neural operator for parametric partial differential equations},
  author    = {Li, Zongyi and Kovachki, Nikola and Azizzadenesheli, Kamyar and Anandkumar, Anima},
  booktitle = {ICLR},
  year      = {2021}
}

@article{cao2023laplacian,
  title   = {Laplacian Neural Operator: Learning stable PDE solvers with implicit layers},
  author  = {Cao, Yunan and Li, Zongyi and Azizzadenesheli, Kamyar and Anandkumar, Anima},
  journal = {NeurIPS},
  year    = {2023}
}

@article{kharazmi2019variational,
  title={Variational physics-informed neural networks for solving partial differential equations},
  author={Kharazmi, Ehsan and Zhang, Zhongqiang and Karniadakis, George Em},
  journal={arXiv preprint arXiv:1912.00873},
  year={2019}
}

@article{kovachki2021neural,
  title={Neural operator: A learning framework for operators},
  author={Kovachki, Nikola and others},
  journal={arXiv:2108.08481},
  year={2021}
}

@article{wang2025kolmogorov,
  title={Kolmogorov--Arnold-Informed neural network: A physics-informed deep learning framework for solving forward and inverse problems based on Kolmogorov--Arnold Networks},
  author={Wang, Yizheng and Sun, Jia and Bai, Jinshuai and Anitescu, Cosmin and Eshaghi, Mohammad Sadegh and Zhuang, Xiaoying and Rabczuk, Timon and Liu, Yinghua},
  journal={Computer Methods in Applied Mechanics and Engineering},
  volume={433},
  pages={117518},
  year={2025},
  publisher={Elsevier}
}

@article{thakolkaran2025can,
  title={Can kan cans? input-convex kolmogorov-arnold networks (kans) as hyperelastic constitutive artificial neural networks (cans)},
  author={Thakolkaran, Prakash and Guo, Yaqi and Saini, Shivam and Peirlinck, Mathias and Alheit, Benjamin and Kumar, Siddhant},
  journal={Computer Methods in Applied Mechanics and Engineering},
  volume={443},
  pages={118089},
  year={2025},
  publisher={Elsevier}
}

@article{han2018solving,
  title={Solving high-dimensional partial differential equations using deep learning},
  author={Han, Jiequn and Jentzen, Arnulf and E, Weinan},
  journal={Proceedings of the National Academy of Sciences},
  volume={115},
  number={34},
  pages={8505--8510},
  year={2018},
  publisher={National Academy of Sciences}
}

@article{weinan2020machine,
  title={Machine learning and computational mathematics},
  author={Weinan, E},
  journal={arXiv preprint arXiv:2009.14596},
  year={2020}
}

@article{karniadakis2021physics,
  title={Physics-informed machine learning},
  author={Karniadakis, George Em and Kevrekidis, Ioannis G and Lu, Lu and Perdikaris, Paris and Wang, Sifan and Yang, Liu},
  journal={Nature Reviews Physics},
  volume={3},
  number={6},
  pages={422--440},
  year={2021},
  publisher={Nature Publishing Group UK London}
}

@article{raissi2019physics,
  title={Physics-informed neural networks: A deep learning framework for solving forward and inverse problems involving nonlinear partial differential equations},
  author={Raissi, Maziar and Perdikaris, Paris and Karniadakis, George E},
  journal={Journal of Computational physics},
  volume={378},
  pages={686--707},
  year={2019},
  publisher={Elsevier}
}

@article{mao2023physics,
  title={Physics-informed neural networks with residual/gradient-based adaptive sampling methods for solving partial differential equations with sharp solutions},
  author={Mao, Zhiping and Meng, Xuhui},
  journal={Applied Mathematics and Mechanics},
  volume={44},
  number={7},
  pages={1069--1084},
  year={2023},
  publisher={Springer}
}

@article{wu2023comprehensive,
  title   = {A comprehensive study of non-adaptive and residual-based adaptive sampling for physics-informed neural networks},
  author  = {Wu, Chenxi and Zhu, Min and Tan, Qinyang and Kartha, Yadhu and Lu, Lu},
  journal = {Computer Methods in Applied Mechanics and Engineering},
  volume  = {403},
  pages   = {115671},
  year    = {2023},
  doi     = {10.1016/j.cma.2022.115671}
}

@article{jagtap2020xpinn,
  title   = {Extended physics-informed neural networks (XPINNs): A generalized space-time domain decomposition based deep learning framework for nonlinear partial differential equations},
  author  = {Jagtap, Ameya D. and Karniadakis, George Em},
  journal = {Communications in Computational Physics},
  volume  = {28},
  number  = {5},
  pages   = {2002--2041},
  year    = {2020}
}

@article{shukla2021parallel,
  title   = {Parallel physics-informed neural networks via domain decomposition},
  author  = {Shukla, Khemraj and Jagtap, Ameya D. and Karniadakis, George Em},
  journal = {Journal of Computational Physics},
  volume  = {447},
  pages   = {110683},
  year    = {2021},
  doi     = {10.1016/j.jcp.2021.110683}
}

@article{wang2021eigenpinn,
  title   = {On the eigenvector bias of Fourier feature networks: From regression to solving multi-scale PDEs with physics-informed neural networks},
  author  = {Wang, Sifan and Wang, Hanwen and Perdikaris, Paris},
  journal = {Computer Methods in Applied Mechanics and Engineering},
  volume  = {384},
  pages   = {113938},
  year    = {2021},
  doi     = {10.1016/j.cma.2021.113938}
}

@article{cai2019mscalednn,
  title   = {Multi-scale deep neural networks for solving high dimensional PDEs},
  author  = {Cai, Wei and Xu, Zhi-Qin John},
  journal = {arXiv preprint arXiv:1910.11710},
  year    = {2019}
}

@article{pang2019fpinn,
  title   = {fPINNs: Fractional physics-informed neural networks},
  author  = {Pang, Guofei and Lu, Lu and Karniadakis, George Em},
  journal = {SIAM Journal on Scientific Computing},
  volume  = {41},
  number  = {4},
  pages   = {A2603--A2626},
  year    = {2019},
  doi     = {10.1137/18M1229845}
}

@article{garg2024_vswno,
  title   = {Neuroscience-inspired neural operator for partial differential equations},
  author  = {Garg, Atul and Mallick, Bobby and Karniadakis, George Em},
  journal = {Journal of Computational Physics},
  volume  = {500},
  pages   = {112809},
  year    = {2024},
  doi     = {10.1016/j.jcp.2024.112809}
}

@article{chen2025_ndo,
  title   = {Neural dynamical operator: Continuous spatial–temporal model with gradient-based and derivative-free optimization methods},
  author  = {Chen, Yiyuan and Lu, Lu and Karniadakis, George Em},
  journal = {Journal of Computational Physics},
  volume  = {514},
  pages   = {113190},
  year    = {2025},
  doi     = {10.1016/j.jcp.2024.113190}
}

@article{kim2025bekan,
  title={BEKAN: Boundary condition-guaranteed evolutionary Kolmogorov-Arnold networks with radial basis functions for solving PDE problems},
  author={Kim, Bongseok and Zhang, Jiahao and Lin, Guang},
  journal={arXiv preprint arXiv:2510.03576},
  year={2025}
}

@article{de2024wpinns,
  title={wPINNs: Weak physics informed neural networks for approximating entropy solutions of hyperbolic conservation laws},
  author={De Ryck, Tim and Mishra, Siddhartha and Molinaro, Roberto},
  journal={SIAM Journal on Numerical Analysis},
  volume={62},
  number={2},
  pages={811--841},
  year={2024},
  publisher={SIAM}
}

@article{xu2021weak,
  title={Weak form theory-guided neural network (TgNN-wf) for deep learning of subsurface single-and two-phase flow},
  author={Xu, Rui and Zhang, Dongxiao and Rong, Miao and Wang, Nanzhe},
  journal={Journal of Computational Physics},
  volume={436},
  pages={110318},
  year={2021},
  publisher={Elsevier}
}

@article{wang2025wf,
  title={WF-PINNs: solving forward and inverse problems of burgers equation with steep gradients using weak-form physics-informed neural networks},
  author={Wang, Xianke and Yi, Shichao and Gu, Huangliang and Xu, Jing and Xu, Wenjie},
  journal={Scientific Reports},
  volume={15},
  number={1},
  pages={40555},
  year={2025},
  publisher={Nature Publishing Group UK London}
}

\end{document}